\documentclass{article}

\usepackage[preprint]{neurips_2026}

\usepackage[T1]{fontenc}
\usepackage[utf8]{inputenc}
\usepackage{microtype}
\usepackage{hyperref}
\usepackage{url}
\usepackage{booktabs}
\usepackage{amsmath}
\usepackage{amssymb}
\usepackage{graphicx}
\usepackage{xcolor}
\usepackage{enumitem}
\usepackage{subcaption}
\usepackage{multirow}
\usepackage{tikz}
\usepackage{listings}

\definecolor{light-gray}{gray}{0.97}

\newcommand{\jplot}[3]{%
	\begin{tikzpicture}[baseline=-0.5ex, x=3.00cm, y=1cm]
		\useasboundingbox (-0.60, -0.20) rectangle (0.20, 0.20);
		\draw[gray!30, thin] (-0.60, 0) -- (0.20, 0);             
		\draw[gray!30, dotted, thick] (0, -0.43) -- (0, 0.43);  
		\draw[very thick] (#1, 0) -- (#2, 0);                    
		\fill (#3, 0) circle (1.5pt);                       
	\end{tikzpicture}%
}

\title{Bias and Uncertainty in LLM-as-a-Judge Estimation}

\author{
James Fiedler \\
Indeed Inc. \\
\texttt{jfiedler@indeed.com}
}

\begin{document}

\maketitle

\begin{abstract}
LLM-as-a-Judge evaluation has become a standard tool for assessing base model performance. However, characterizing performance via the naive estimator, i.e., raw judge outputs, is systematically biased. Recent work has proposed estimators to correct this bias, but their reliability depends critically on judge quality and, for model comparisons, on calibration stability. Sharing calibration across compared models is practically attractive but can introduce severe bias, including cases where the comparison estimate points in the wrong direction with high apparent confidence. We study these failure modes through analytical results, simulations over judge quality ($J$) and cross-model calibration instability ($\Delta J$), and a real-data MMLU-Pro case study with sign reversal. We propose $J$ and $\Delta J$ as diagnostics for when corrected estimates, especially shared-calibration comparisons, are likely unreliable, and provide reporting guidance for LaaJ evaluation.
\end{abstract}

\section{Introduction} \label{sec:intro}

LLM-as-a-Judge (LaaJ) evaluation has become a standard tool for assessing base model performance across a growing range of tasks and benchmarks \citep{zheng-laaj,liu-etal-2023-g,dubois2024length}. Raw judge outputs are often treated as direct measurements of model quality; reporting these outputs without correction or confidence intervals has become common practice \citep{lee2026correctly}.

The core statistical problem is not only bias, but whether corrected judge outputs can be trusted as estimates of true model performance. Bias-corrected estimators \citep{lee2026correctly,chen2026efficient} can substantially improve estimates, but they are not automatically trustworthy. Their reliability depends on judge quality, summarized by Youden's $J$, and on whether the calibration assumptions required to justify the correction actually hold. The central issue here is not estimation efficiency, but whether the resulting estimate and uncertainty statement correspond to the intended quantity. We therefore treat LaaJ estimation as an evaluation method that should itself be stress-tested before it is used to support scientific claims.

\citet{lee2026correctly} show that the Rogan-Gladen estimator (RG) can remain valid under distribution shift when judge predictions depend only on latent correctness rather than, for example, properties of the base model being evaluated. This invites a practical strategy of reusing calibration estimates across multiple evaluation contexts, to save on labeling time and cost. However, such reuse is only justified if judge calibration is effectively invariant across those contexts. When it is not, estimation can be severely distorted. For model comparison, the distortion can be large enough to reverse the sign of the estimate, producing an apparently confident conclusion that points in the wrong direction.

This paper studies LaaJ estimation for base-model assessment, covering both single-model accuracy estimation and model comparison, with particular attention to comparison because it is more fragile. Our analytical and simulation results focus on the case where the judge outputs a binary label; it is the most analytically transparent setting and is one of, if not the, most common in practice. However, the core lessons apply more broadly, including to ordinal and continuous judge outputs.

We study these failure modes through three lines of evidence. Analytical results characterize the bias structure under shared calibration and the role of $J$ in determining estimator stability, where $J = \mathrm{sensitivity} + \mathrm{specificity} - 1$ is Youden's index \citep{peirce1884numerical,youden1950index}. Simulations demonstrate the impact that judge quality ($J$) and cross-model calibration instability ($\Delta J$) can have on bias and interval coverage. We follow up with a real-data case study using two subjects from the MMLU-Pro benchmark \citep{wang2024mmlu} that corroborates and expands on the simulation. In one subject estimators perform relatively well. The other is more severe and reveals additional failures, including a sign-reversal failure, where a shared-calibration comparison yields a confidently negative estimate for a truly positive accuracy difference.

Our contributions are:
\begin{enumerate}[leftmargin=*, itemsep=0.3em]
	\item We characterize the fragility of bias-corrected LaaJ estimators, showing that estimates for both single-model quality and model comparison can degrade sharply as judge quality ($J$) falls, and that the severity and form of failure differ across estimators.
	\item We analyze shared calibration as a risk factor in both single-model and comparison estimation, showing that it induces $1/J$ sensitivity to calibration mismatch and, for RG, can yield severely amplified systematic bias when $J$ is small.
	\item We propose $J$ and $\Delta J$ as prerequisite diagnostics that should be checked and reported before trusting corrected LaaJ estimates, with increasing caution with decreasing $J$ or increasing $\Delta J$.
	\item We demonstrate estimator failure modes through simulation and a real-data MMLU-Pro case study, and provide practical reporting guidance for LaaJ evaluation.
\end{enumerate}

\section{Related work} \label{sec:related-work}

LLM-as-a-Judge (LaaJ) evaluation has been widely adopted since MT-Bench \citep{zheng-laaj} and G-Eval \citep{liu-etal-2023-g} helped establish the use of LLMs to grade model responses at scale. Subsequent work has extended LaaJ to preference-style evaluation \citep{dubois2024length} and created structured judge benchmarks \citep{tan2025judgebench}. Across this literature, naive judge scores are routinely treated as estimates of base-model quality without calibration correction or confidence intervals \citep{zheng-laaj,tan2025judgebench,lee2026correctly}. \citet{miller2024adding} is a strong precedent for statistical uncertainty reporting in LLM evaluation more broadly. Our focus is on calibration challenges specific to LaaJ that make naive uncertainty statements unreliable even after accounting for sampling variation.

\citet{lee2026correctly} showed that naive LaaJ estimates are biased and proposed applying the Rogan-Gladen estimator \citep{rogan1978estimating} to correct for judge sensitivity and specificity. They establish that the RG estimator remains valid under test-set distribution shift when the judge's predictions depend only on latent correctness, and validate on Chatbot Arena win-rate data \citep{chiang2024chatbotarenaopenplatform}. \citet{chen2026efficient} subsequently compared RG, PPI\texttt{++} \citep{angelopoulos2024ppiefficient}, and an EIF-based estimator, showing that PPI\texttt{++} and the EIF estimator are more efficient than RG in the binary case. Our focus differs: rather than efficiency, we study when corrected estimates break down, especially in model comparison under shared calibration and cross-model instability. Neither paper studies shared calibration, proposes diagnostics for estimator trustworthiness, or foregrounds the structural distinction between judge-centric shared calibration and model-specific calibration that shapes our analysis.

The epidemiological literature has long recognized that the Rogan-Gladen estimator should not be applied indiscriminately. \citet{greiner2003decision} showed via second-order Taylor-series analysis, confirmed by simulation, that the corrected estimate can have higher mean squared error than the naive estimate. \citet{lang2014confidence} derived adjusted confidence intervals accounting for uncertainty in sensitivity and specificity, establishing that the variance of the RG estimator scales as $1/J^2$. This instability is intrinsic to the RG formula but not to the estimation problem itself: PPI\texttt{++} degrades more gracefully as judge quality falls. Both papers address single-population estimation. We extend to model comparison, where shared calibration can introduce systematic bias when judge calibration differs across models, a failure mode we are not aware of prior work characterizing.

Within LLM evaluation literature, \citet{chen2026efficient} find in simulations that the efficiency advantage of PPI\texttt{++} and their EIF-based estimator over RG is most pronounced at low judge quality, and that RG can overcover when judge quality and labeled fraction are both small. Their analysis focuses on efficiency under correctly specified assumptions; they do not study shared calibration, $\Delta J$, or conditions under which corrected estimates become systematically biased or yield wrong-signed comparisons. \citet{collot-etal-2026-balanced} argue that Youden's $J$ is the right metric for selecting LLM judges, grounding this in its prevalence-independence properties. We share their view, but our use of $J$ is complementary: rather than selecting judges, we focus on bias-corrected estimation, asking whether a chosen judge's estimates are reliable for both single-model and model-comparison evaluations. The cross-model stability measure $\Delta J$ is, to our knowledge, novel as a formal diagnostic criterion.

\section{Naive and Bias-Corrected Estimation} \label{sec:naive-and-corrected-est}

\subsection{Setup} \label{subsec:naive-bias-setup}

In LaaJ evaluation, the judge's labels are themselves noisy \citep{zheng-laaj,wang-etal-2024-large-language-models-fair}. That creates two distinct inferential problems. First, raw judge estimates of performance are generally a biased estimate. Second, model comparison is more fragile than single-model estimation because calibration assumptions used to debias the judge may or may not hold equally well across models.

LLM-as-a-Judge evaluation takes two common forms. In preference evaluation, the judge is presented multiple model outputs together and asked which is better; the estimand is the win rate $\Pr(\text{judge prefers } A)$. In accuracy/quality evaluation, the judge grades each response independently against a rubric or answer key; the estimand is, e.g., true accuracy $\theta = \Pr(Z = 1)$. Preference evaluation is natural when direct comparison is the primary goal; accuracy evaluation is natural when single-model performance is itself of interest. Both forms require human-labeled calibration data to estimate and correct for judge error. Accuracy evaluation extends naturally to model comparison as $\Delta\theta = \theta_A - \theta_B$. Many LaaJ pipelines are built for accuracy evaluation and reused for model comparison, making $\Delta\theta$ the natural downstream estimand.

We focus on the common binary setting in which the judge emits a label $\hat Z \in \{0,1\}$ for whether a model response is correct. Let $Z \in \{0,1\}$ denote the corresponding ground-truth label. The target quantity is the model's true accuracy $\theta = \Pr(Z = 1)$. Although we focus on the binary case, the core lessons extend more broadly to non-binary outputs.

\subsection{The naive estimator is biased}

The naive estimator for base-model performance is the fraction of responses judged as correct,
\[
\hat p = \frac{1}{n} \sum_{i=1}^n \hat z_i.
\]
But $\hat p$ estimates $p = \Pr(\hat Z = 1)$, not $\theta$. If the judge has sensitivity $q_1 = \Pr(\hat Z = 1 \mid Z = 1)$ and specificity $q_0 = \Pr(\hat Z = 0 \mid Z = 0)$, then
\begin{equation}
	p = \theta \cdot q_1 + (1-\theta)(1-q_0),
	\label{eq:measurement-model}
\end{equation}
so $\hat p$ is unbiased for $\theta$ only when the judge is perfect.

In practice, judges are variable and far from perfect \citep{tan2025judgebench}. Judge inaccuracy remains even with prompt optimization \citep{zheng-laaj,xie2025an} and fine-tuning to match human labels \citep{wang2024pandalm,dubois2023alpacafarm,huang-etal-2025-empirical}.

The same bias structure applies to binary win-rate judges. If option A is preferred by humans with probability $\pi = \Pr(Z = 1)$, then Equation~\eqref{eq:measurement-model} holds with $\theta$ replaced by $\pi$: the naive win rate estimates the judge's marginal preference probability, not the true human preference rate.

\subsection{Bias-corrected estimators}

To correct the naive estimator's bias, we use a calibration set of pairs $(Z_j, \hat Z_j)$ where the true label is known from human annotation. The Rogan-Gladen estimator (RG; \citep{rogan1978estimating}; \citep{lee2026correctly}) uses estimated sensitivity and specificity from that calibration set:
\[
\hat\theta_{\mathrm{RG}} = \frac{\hat p + \hat q_0 - 1}{\hat q_0 + \hat q_1 - 1}.
\]
The denominator is Youden's $J$, a standard summary of test quality. When $\hat J$ is small, the correction becomes unstable and interval width can increase sharply.

We also consider PPI\texttt{++} \citep{angelopoulos2024ppiefficient}, a refinement of the PPI estimator \citep{angelopoulos2023ppi}, which in this setting can be written as
\[
\hat{\theta}_{\mathrm{PPI}\texttt{++}} = \bar Y_\mathrm{cal} + \lambda \left( \hat p_{\mathrm{test}} - \hat p_{\mathrm{cal}} \right),
\]
with $\bar Y_\mathrm{cal}$, $\hat p_{\mathrm{cal}}$ determined from the calibration set and $\lambda$ chosen to reduce variance. \citet{chen2026efficient} introduced an EIF-based estimator shown to be asymptotically equivalent to PPI\texttt{++} in the binary case.

There is an important structural distinction in how these estimators use calibration data. RG calibration is independent of any base model and can in principle be reused across base models under a judge-invariance assumption. PPI\texttt{++} calibration depends on responses from the particular base model being evaluated. This distinction shapes the shared-calibration analysis in the following section and is described more fully as ``judge-centric'' and ``model-centric'' in Appendix~\ref{appendix:rg-vs-ppi-calibration}.

\section{Shared Calibration and Diagnostics}

\subsection{Re-use of calibration sets}

\citet{lee2026correctly} show that the RG estimator can remain valid under test-set distribution shift if the judge's predictions depend only on latent correctness, and no dependence, for example, on the base model. This invites a practical shortcut: estimate judge calibration once and reuse it across base models or test sets. Practitioner guidance and engineering studies already recommend rerunning LaaJ evaluations on the same human-labeled dataset when the judge changes \citep{langchain-frozen-cali,microsoft-trusted-judges}. Our results address reuse in a different direction: across base models rather than across judges. A similar cross-model stability assumption also appears in \citet{collot-etal-2026-balanced}, who study judge error rates across assistant models.

This shortcut is only justified if judge calibration is effectively invariant across the models being evaluated. We study how robust this strategy is in practice and show that even moderate departures from calibration invariance can materially distort estimates, for single-model accuracy and, more severely, for model comparisons.

\subsection{Why model comparison is harder}

For a single model, we need to correct the judge's systematic bias and propagate uncertainty from both the test set and the calibration set. For model comparison, we must also pair models on test items and consider whether calibration information can legitimately be shared across models.

RG can reuse calibration estimates if the judge's $q_0$ and $q_1$ are stable across models, which can substantially reduce labeling cost. PPI\texttt{++}, by contrast, generally requires model-specific calibration data (we explore this a bit further in Appendix~\ref{appendix:ppi-shared-cali}). But RG's labeling-cost advantage depends entirely on the judge-invariance assumption holding. If it does not, a cross-model RG comparison will have a bias that scales with $1/J$, and this $1/J$ amplification is forced on any estimator using shared calibration, not just RG (Appendix~\ref{appendix:rg-vs-ppi-calibration}). Appendix~\ref{appendix:rg-variance-estimator} derives the explicit variance formula, showing how $1/J^2$ scaling enters and why pairing test-set items matters for comparison estimates.

\subsection{Uncertainty estimation and the role of the bootstrap}

Confidence intervals (CIs) must propagate uncertainty from both the test set and the calibration set, and in model comparisons, the dependence induced by paired items and shared calibration. We use the bootstrap throughout, resampling at the unit that matches the evaluation design: separately resampling calibration and test items, and resampling paired item indices for model comparisons. This single procedure covers raw differences, percentage changes, and shared or model-specific calibration without requiring a separate variance derivation for each case.

\subsection{Diagnosing when comparison estimates can be trusted}

Before asking which estimator is most efficient, we must ask whether calibration assumptions are plausible and whether estimates are stable. We therefore distinguish two questions:
\begin{enumerate}[leftmargin=*, itemsep=0.3em]
	\item Is judge quality $J$ large enough to produce stable corrected estimates?
	\item For shared-calibration comparisons: is $\widehat{\Delta J}$ small relative to its uncertainty?
\end{enumerate}

We assess both questions using per-model calibration diagnostics, reporting point estimates and confidence intervals. For the second diagnostic, the farther $\widehat{\Delta J}$ is from zero, the less we can trust the calibration-invariance assumption. Inclusion of zero in the confidence interval for $\Delta J$ is a necessary, but not sufficient, condition for shared calibration to be defensible.

The first diagnostic matters regardless of estimator. Low $\widehat{J}$ causes unstable estimates under both RG and, to a lesser extent, PPI\texttt{++}, and with low $\widehat{J}$ the right response is often not ``switch estimators and proceed'' but ``treat all estimates with more caution.'' Low judge quality is not a pathological edge case: \citet{tan2025judgebench} document judge accuracy spanning from near-chance to around 80\%. Near-tie comparisons remain fragile even when $\widehat{\Delta J}$ looks small, because modest residual bias can still create a spuriously directional estimate.

For RG, these diagnostic requirements have a practical implication: a portion of the potential labeling-efficiency gains are unrealizable in many settings, because per-model calibration data is needed to assess whether shared calibration is justified. Once stability is established for a particular judge and set of base models, practitioners can benefit from shared calibration for subsequent evaluations.

\section{Simulations over $J$ and $\Delta J$} \label{sec:simulations}

The simulation space is too rich to exhaust in one section, so we focus on a small set of mechanisms most relevant to judge-based model comparison: overall judge quality, calibration instability across models, and their interaction with shared versus model-specific correction. Section~\ref{sec:mmlu-pro-judge} complements these controlled simulations with a real-data case study.

\begin{figure}[t]
	\centering
	\begin{subfigure}[b]{0.49\textwidth}
		\centering
		\includegraphics[width=\linewidth]{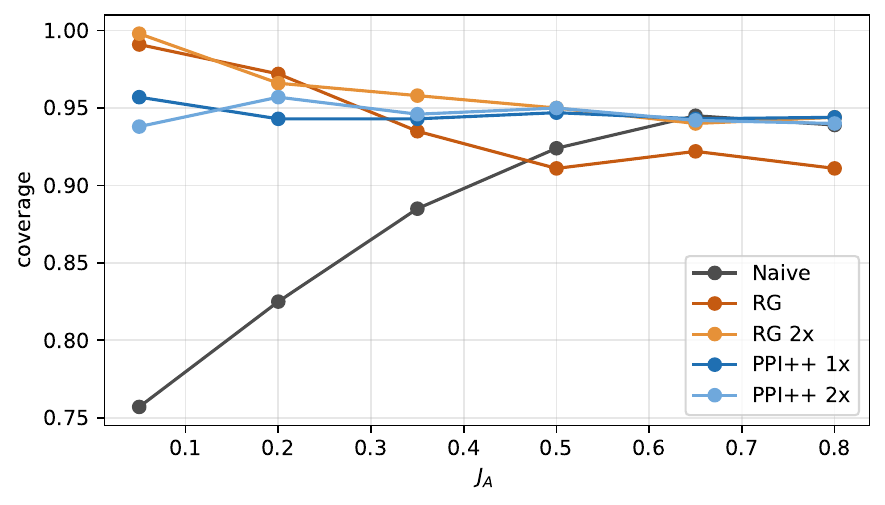}
		\caption{Coverage vs.\ $J_A$ ($\Delta J = 0.05$ fixed)}
		\label{fig:sim-coverage-J}
	\end{subfigure}
	\hfill
	\begin{subfigure}[b]{0.49\textwidth}
		\centering
		\includegraphics[width=\linewidth]{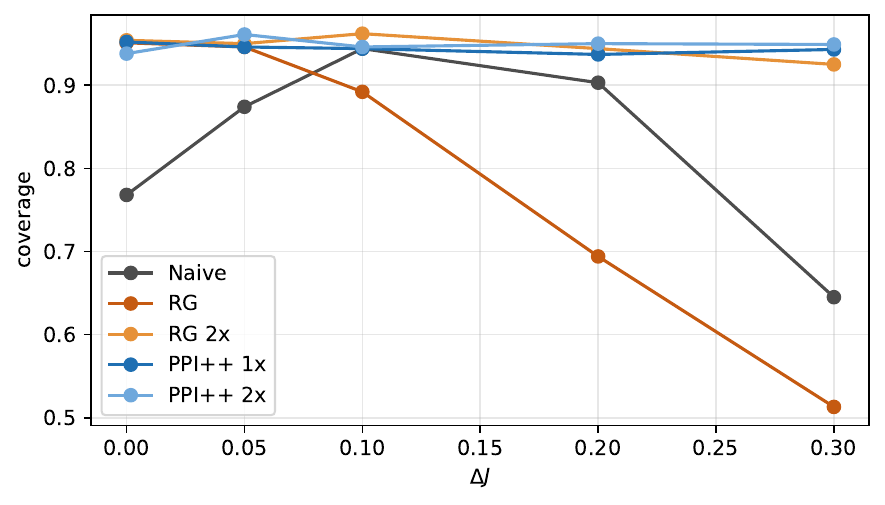}
		\caption{Coverage vs.\ $\Delta J$ ($J_A = 0.3$ fixed)}
		\label{fig:sim-coverage-dJ}
	\end{subfigure}

	\vspace{0.5em}

	\begin{subfigure}[b]{0.49\textwidth}
		\centering
		\includegraphics[width=\linewidth]{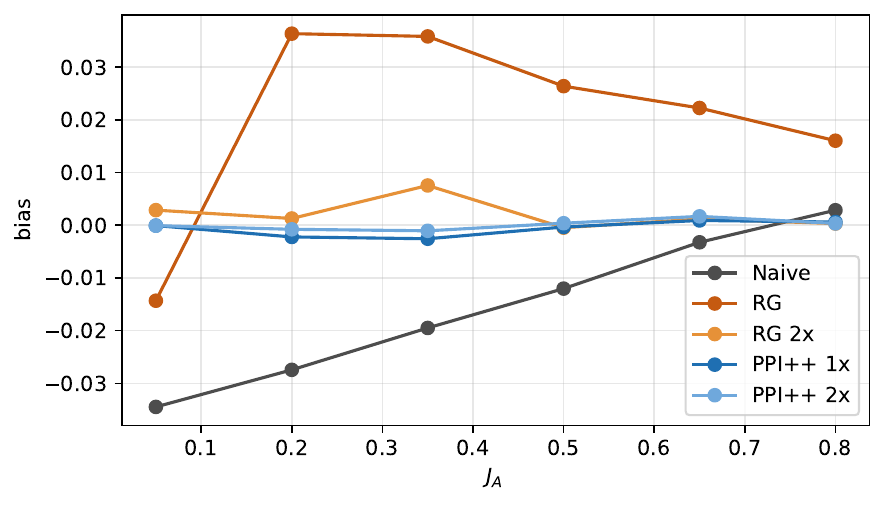}
		\caption{Bias vs.\ $J_A$ ($\Delta J = 0.05$ fixed)}
		\label{fig:sim-bias-J}
	\end{subfigure}
	\hfill
	\begin{subfigure}[b]{0.49\textwidth}
		\centering
		\includegraphics[width=\linewidth]{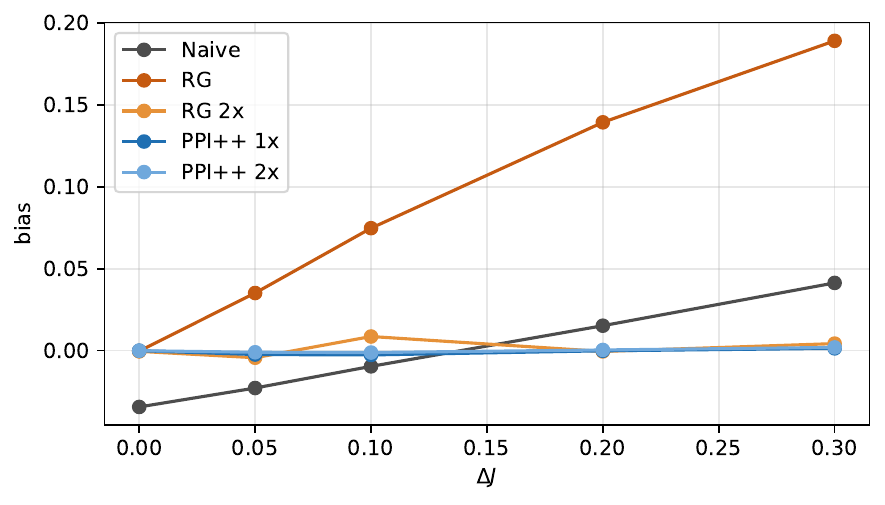}
		\caption{Bias vs.\ $\Delta J$ ($J_A = 0.3$ fixed)}
		\label{fig:sim-bias-dJ}
	\end{subfigure}

	\caption{Simulation results under $q_0 = q_1$ with true comparison effect $\delta = 0.05$. Dotted horizontal line in coverage panels marks nominal 95\%. Left column sweeps $J_A$ with $\Delta J = 0.05$ fixed; right column sweeps $\Delta J$ with $J_A = 0.3$ fixed.}
	\label{fig:judge-sim}
\end{figure}

We simulate binary LaaJ model comparison to isolate how estimator behavior changes with judge quality $J$ and cross-model calibration instability $\Delta J$. In this simulation section, the true accuracy gap is parameterized as $\theta_B - \theta_A = 0.05$, and the calibration gap as $\Delta J = J_B - J_A$. We use equal specificity $q_0$ and sensitivity $q_1$ for analytical transparency, and run two sweeps: one varying $J_A$ with $\Delta J = 0.05$ fixed, and one varying $\Delta J$ with $J_A = 0.3$ fixed. The $\Delta J$ sweep therefore considers nonnegative values in which model $B$ is increasingly easier for the judge than model $A$. These conditions are favorable to model-specific correction with both the measurement model and the estimator assumptions well-matched. Still, shared-calibration RG fails materially here, and real-data failures can be harsher still. Full simulation parameters are in the accompanying code. Figure~\ref{fig:judge-sim} presents our results.

\paragraph{Varying $J_A$.}
As judge quality falls, estimator performance deteriorates, but in different ways. The naive estimator is biased throughout, but the bias is small at the highest $J$ values. RG undercovers even at moderate $J$ because the fixed calibration gap, though small, induces a persistent shared-calibration bias; at low $J$, $1/J$ amplification makes this worse. RG\,$2\times$ removes the shared-calibration assumption and avoids much of RG's systematic bias, but PPI\texttt{++} remains the better-behaved estimator family in this simulation. PPI\texttt{++} stays near nominal coverage across the full range, and PPI\texttt{++}\,$2\times$ provides a tighter version of the same behavior. The lesson is not simply that removing shared calibration solves the problem: model-specific correction can still be uncompetitive when variance is high.

\paragraph{Varying $\Delta J$.}
At $\Delta J = 0$, the naive estimator is already biased because $J_A = 0.3$ is too low for raw judge outputs to recover the true comparison. As the calibration gap increases, shared-calibration RG bias grows systematically; by $\Delta J = 0.3$, RG has both higher absolute bias and lower coverage than the naive estimator, so calibration correction actively makes the comparison estimate worse. PPI\texttt{++} remains comparatively stable across the full sweep because model-specific calibration insulates it from cross-model mismatch. The central mechanism is that shared calibration routes any calibration mismatch directly into comparison-estimation bias, amplified by $1/J$.

\section{MMLU-Pro case study} \label{sec:mmlu-pro-judge}

We apply these ideas to MMLU-Pro using two subjects with very different judge behavior: math and biology. This pair contrasts a relatively well-behaved regime with a failure regime. The compared base models are Gemma\,3\,12B and Qwen\,2.5\,7B, and we use two judges, Mistral Large and Gemma\,4\,31B. We report diagnostic estimates in Table~\ref{tab:judge-J}, comparison estimates in Table~\ref{tab:judge-theta-diffs}, and per-model accuracy estimates in Figure~\ref{fig:judge-cis}. Full setup details are in Appendix~\ref{appendix:mmlu-pro-laaj}. The intervals reflect uncertainty from the realized calibration/test partition, finite calibration size, and, where applicable, repeated judge runs or estimator-specific calibration subsampling; generalizing beyond these specific MMLU-Pro questions requires assuming the test items are representative of a broader target population.

Additional subjects in Appendix~\ref{appendix:more-laaj-results} show qualitatively similar patterns with more mixed severity.

\begin{table}[t]
	\centering
	\resizebox{\textwidth}{!}{
		\begin{tabular}{llcccc}
			\toprule
			Subject & Judge & $\hat{J}_{\mathrm{Gemma\,3}}$ & $\hat{J}_{\mathrm{Qwen\,2.5}}$ & $\widehat{\Delta J}$ & $\widehat{\Delta J}$ plotted
			\\
			\midrule
			\multirow{4}{*}{Math}
			& \multirow{2}{*}{Mistral Large}
			& $0.397$ & $0.516$ & $-0.119$
			& \multirow{2}{*}{\jplot{-0.253}{0.016}{-0.119}} \\
			& & \small$[0.297,\ 0.494]$
			& \small$[0.418,\ 0.612]$
			& \small$[{-0.253},\ 0.016]$ & \\[4pt]
			& \multirow{2}{*}{Gemma\,4\,31B}
			& $0.642$ & $0.671$ & $-0.030$
			& \multirow{2}{*}{\jplot{-0.140}{0.082}{-0.030}} \\
			& & \small$[0.551,\ 0.728]$
			& \small$[0.582,\ 0.759]$
			& \small$[{-0.140},\ 0.082]$ & \\
			\midrule
			\multirow{4}{*}{Biology}
			& \multirow{2}{*}{Mistral Large}
			& $0.094$ & $0.382$ & $-0.289$
			& \multirow{2}{*}{\jplot{-0.456}{-0.125}{-0.289}} \\
			& & \small$[{-0.056},\ 0.247]$
			& \small$[0.276,\ 0.494]$
			& \small$[{-0.456},\ {-0.125}]$ & \\[4pt]
			& \multirow{2}{*}{Gemma\,4\,31B}
			& $0.181$ & $0.528$ & $-0.347$
			& \multirow{2}{*}{\jplot{-0.505}{-0.190}{-0.347}} \\
			& & \small$[0.027,\ 0.340]$
			& \small$[0.415,\ 0.639]$
			& \small$[{-0.505},\ {-0.190}]$ & \\
			\bottomrule
		\end{tabular}
	}
	\caption{Per-model Youden's $\hat{J} = \hat{q}_0 + \hat{q}_1 - 1$ and difference
		$\widehat{\Delta J}$ with 95\% bootstrap CIs. Right column shows $\widehat{\Delta J}$ on a common scale $[-0.6,\ 0.2]$; dotted line marks zero. A CI far from zero indicates the shared-calibration assumption is not defensible. Per-model $\hat{q}_0$ and $\hat{q}_1$ are in Table~\ref{tab:judge-q0q1}.}
	\label{tab:judge-J}
\end{table}

Calibration labels play two distinct roles in this case study. First, they determine the labeling cost of each estimator. For a two-model comparison, shared calibration requires labels for only one calibration set, while model-specific calibration requires a separate set per model, doubling the annotation cost. To make that tradeoff visible, we include two variants of RG and PPI\texttt{++} with different calibration budgets.

Second, calibration labels serve a different purpose for diagnostics than for estimation. Accuracy estimates use only the labels available under each estimator's own calibration design, while the judge-quality diagnostics ($J$ and $\Delta J$) are computed using pooled calibration data from both models. The reason is that the diagnostic question is different: not how accurately a particular estimator recovers model performance under its own labeling budget, but how reliably the judge evaluates responses from each model. In practice, we recommend estimating judge quality in advance, across the relevant judges and base models, using at least part of the available calibration data before committing to a shared-calibration design.

Table~\ref{tab:judge-theta-diffs} reports comparison estimates for five estimators: naive, RG (shared calibration), RG\,$2\times$ (model-specific RG), PPI\texttt{++} (model-specific), and PPI\texttt{++}\,$2\times$ (model-specific, double calibration budget). Calibration requires human-labeled examples to characterize judge error rates. The $2\times$ variants allocate twice as many such labels as the $1\times$ variants, at (presumably) additional annotation cost; comparing them shows the effect of greater calibration investment on estimation quality. Per-model accuracy estimates are shown in Figure~\ref{fig:judge-cis}.

\begin{table}[t]
	\centering
	\small
	\setlength{\tabcolsep}{4pt}
	\resizebox{\textwidth}{!}{
		\begin{tabular}{ll c ccccc}
			\toprule
			Subject & Judge & $\Delta\theta$
			& $\widehat{\Delta\theta}_{\mathrm{naive}}$
			& $\widehat{\Delta\theta}_{\mathrm{RG}}$
			& $\widehat{\Delta\theta}_{\mathrm{RG,2\times}}$
			& $\widehat{\Delta\theta}_{\mathrm{PPI{+}{+}}}$
			& $\widehat{\Delta\theta}_{\mathrm{PPI{+}{+},2\times}}$ \\
			\midrule
			\multirow{4}{*}{Math}
			& \multirow{2}{*}{Mistral Large} & \multirow{2}{*}{$+$0.003}
			& $-$0.046 & $-$0.089 & $-$0.037 & $-$0.009 & $-$0.007 \\
			& & & \footnotesize$[-0.078,\;-0.016]$
			& \footnotesize$[-0.156,\;-0.030]$
			& \footnotesize$[-0.176,\;+0.081]$
			& \footnotesize$[-0.084,\;+0.059]$
			& \footnotesize$[-0.050,\;+0.034]$ \\[4pt]
			& \multirow{2}{*}{Gemma\,4\,31B} & \multirow{2}{*}{$+$0.003}
			& $-$0.038 & $-$0.057 & $-$0.008 & $-$0.008 & $-$0.004 \\
			& & & \footnotesize$[-0.070,\;-0.007]$
			& \footnotesize$[-0.106,\;-0.010]$
			& \footnotesize$[-0.086,\;+0.072]$
			& \footnotesize$[-0.083,\;+0.060]$
			& \footnotesize$[-0.044,\;+0.037]$ \\
			\midrule
			\multirow{4}{*}{Biology}
			& \multirow{2}{*}{Mistral Large} & \multirow{2}{*}{$+$0.048}
			& $-$0.127 & $-$0.330 & $-$0.002 & $+$0.149 & $+$0.142 \\
			& & & \footnotesize$[-0.173,\;-0.079]$
			& \footnotesize$[-0.493,\;-0.194]$
			& \footnotesize$[-0.635,\;+0.528]$
			& \footnotesize$[+0.059,\;+0.237]$
			& \footnotesize$[+0.090,\;+0.195]$ \\[4pt]
			& \multirow{2}{*}{Gemma\,4\,31B} & \multirow{2}{*}{$+$0.048}
			& $-$0.115 & $-$0.218 & $-$0.066 & $+$0.137 & $+$0.133 \\
			& & & \footnotesize$[-0.165,\;-0.063]$
			& \footnotesize$[-0.341,\;-0.117]$
			& \footnotesize$[-0.641,\;+0.346]$
			& \footnotesize$[+0.042,\;+0.224]$
			& \footnotesize$[+0.082,\;+0.185]$ \\
			\bottomrule
		\end{tabular}
	}
	\caption{Gemma\,3 minus Qwen\,2.5 accuracy on math and biology. $\Delta\theta$ is the true difference. Estimates are bootstrap medians with 95\% bootstrap CIs. RG uses shared calibration; RG\,$2\times$ uses model-specific calibration with the RG formula; PPI\texttt{++} variants use model-specific calibration.}
	\label{tab:judge-theta-diffs}
\end{table}

\begin{figure}[t]
	\centering
	\begin{subfigure}[t]{0.500\textwidth}
		\centering
		\includegraphics[width=\linewidth]{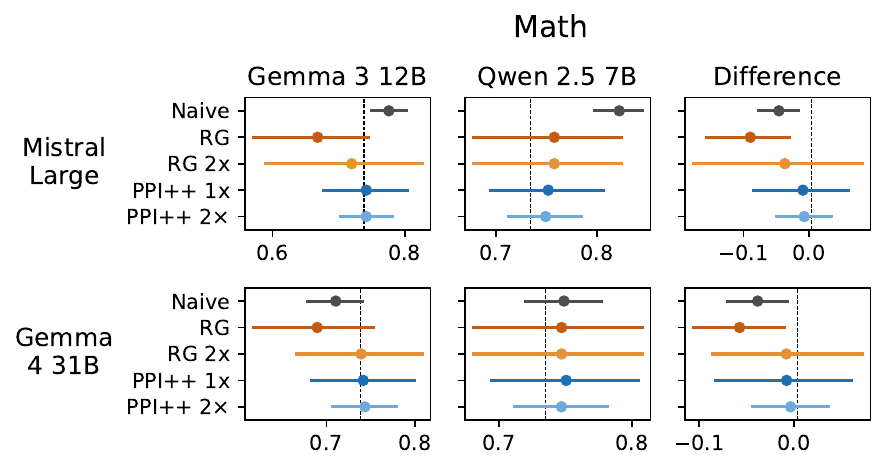}
	\end{subfigure}
	\begin{subfigure}[t]{0.490\textwidth}
		\centering
		\includegraphics[width=\linewidth]{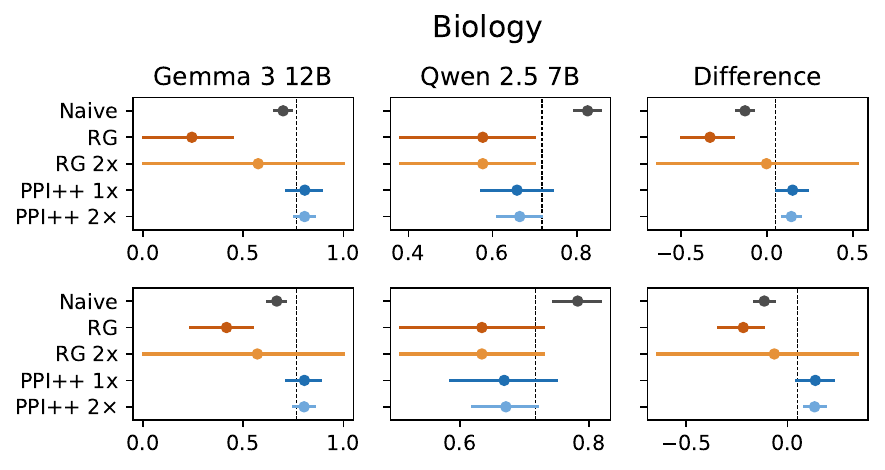}
	\end{subfigure}
	\caption{Bootstrap medians and 95\% confidence intervals for per-model accuracy $\hat\theta$ for the naive estimator and four corrected estimators, under each judge. The dotted vertical line shows true accuracy (mean on the test set). Full numerical results appear in Table~\ref{tab:judge-theta-estimates}.}
	\label{fig:judge-cis}
\end{figure}

\subsection{Math: a relatively well-behaved comparison}

The base models achieve nearly identical accuracy on math (true $\Delta\theta = +0.003$), making this a near-tie regime where any non-trivial comparison estimate is suspect. The diagnostics in Table~\ref{tab:judge-J} look relatively favorable: $\widehat{J}$ is positive and moderate for both models and both judges, and the $\widehat{\Delta J}$ intervals include zero.

Individual per-model estimates (Figure~\ref{fig:judge-cis}) show that the naive estimator is visibly off the true accuracy for both models under both judges. These per-model errors do not cancel in the comparison: the naive comparison yields $-0.046$ and $-0.038$ (CIs excluding zero for both judges) despite a practically zero true difference. Even in this relatively well-behaved regime, correction is therefore necessary.

RG also gives confidently negative comparison estimates ($-0.089$ and $-0.057$, CIs excluding zero): the shared-calibration formula amplifies the per-model error, and Appendix~\ref{appendix:rg-shared-cali-bias} derives a formula that explains the observed bias for Gemma\,3. RG\,$2\times$ (model-specific RG) has medians closer to zero and CIs that include the true value, but the intervals are substantially wider than those of PPI\texttt{++}.

This case therefore shows that shared calibration remains risky even when diagnostics look favorable: in a near-tie regime, it can convert modest per-model bias into a spuriously directional comparison result. Model-specific PPI\texttt{++} is the best-behaved estimator family among those we study in this regime, combining near-zero medians with comparatively tight intervals containing the true difference for both budget levels and both judges.

\subsection{Biology: calibration instability and low judge quality}

The biology diagnostics clearly indicate trouble. The $\widehat{\Delta J}$ intervals decisively exclude zero for both judges, and $\widehat{J}$ for Gemma\,3 is near zero. Both judges are substantially better at evaluating Qwen\,2.5\,7B than Gemma\,3\,12B, with especially low specificity for Gemma\,3 (Appendix, Table~\ref{tab:judge-q0q1}). The diagnostics alone warrant skepticism about any estimate.

Per-model estimates (Figure~\ref{fig:judge-cis}) show that naive systematically underestimates Gemma\,3 and overestimates Qwen\,2.5. Both biases push the comparison in the same wrong direction. PPI\texttt{++} gives the most accurate per-model estimates overall, with intervals that track the true accuracies more closely than the alternatives. The remaining estimators produce substantially biased per-model results; for Gemma\,3 specifically, the near-zero $J$ causes the $1/J$ amplification in RG to produce near-degenerate intervals spanning most of $[0, 1]$.

For model comparison, the true difference is $+0.048$. The naive and RG estimators give confidently wrong-signed estimates with CIs firmly excluding zero. RG\,$2\times$ avoids the confident wrong sign but produces CIs too wide to be useful. PPI\texttt{++} recovers the correct direction, but its estimates are upward-biased and most CIs exclude the true difference from above.

Biology is the sharpest failure mode in the paper: diagnostics give clear advance warning, multiple estimators report confidently wrong-signed comparisons, and even the most robust option here, PPI\texttt{++}, has coverage issues. Practitioners should have serious doubt about any estimator when judge quality is this poor.

\subsection{Main practical lesson}

Naive estimation fails in every comparison, across both subjects and both judges, establishing that correction is necessary. But correction is not sufficient. In math, RG fails despite favorable diagnostics, because shared calibration introduces excess error even when $\widehat{\Delta J}$ is small. In biology, no estimator fully succeeds: PPI\texttt{++} performs far better than the alternatives considered here, but its comparison intervals do not always contain the true value when judge quality is very low.

For per-model estimates, the naive estimator is also biased in both subjects. This supports the analytical and simulation results that raw judge scores should not be treated as unbiased performance estimates regardless of whether a comparison is being made. We also saw that shared calibration can bias per-model estimates, as explained in Appendix~\ref{appendix:rg-shared-cali-bias}.

Our recommended diagnostics $J$ and $\Delta J$ provide warning signals rather than acceptance criteria. When per-model calibration is feasible, judge quality is moderate to high for both models, and $\widehat{\Delta J}$ is small with a CI near zero, model-specific PPI\texttt{++} is the most consistently well-behaved choice among the estimators we study. When the diagnostics indicate trouble, as with the biology use case, all estimates should be treated with caution.

\section{Practical guidance for LaaJ evaluation}

Raw judge scores are not unbiased estimates of model accuracy, and the reliability of corrected estimators depends on judge quality and calibration design. The following checklist summarizes what LaaJ evaluation reports should make explicit.

\begin{enumerate}[leftmargin=*, itemsep=0.3em]
	\item \textbf{State the estimand and whether it is corrected.} Report whether results use raw outputs or are bias-corrected. If bias-corrected, name the estimator and describe the calibration design.
	
	\item \textbf{Distinguish single-model estimation from model comparison.} The two tasks have different calibration requirements, different error propagation, and different diagnostic requirements.
	
	\item \textbf{Report calibration design explicitly.} State whether calibration is shared across models or model-specific, and report calibration set size. Shared calibration requires justification; it cannot be a default design choice.
	
	\item \textbf{Report confidence intervals, and state what randomness they represent.} Clarify whether the intervals reflect calibration sampling, test-set sampling, repeated judge runs, estimator-specific subsampling, or some combination. For model comparisons, preserve pairing when resampling.
	
	\item \textbf{Report judge-quality diagnostics: $\hat{q}_0$, $\hat{q}_1$, and $\hat{J}$.} These are necessary for interpreting corrected estimates. Include confidence intervals.
	
	\item \textbf{For model comparisons, also report $\widehat{\Delta J}$ with a confidence interval.} A large $|\widehat{\Delta J}|$ or a confidence interval excluding zero is a strong warning sign that shared-calibration assumptions are not defensible.
	
	\item \textbf{When diagnostics indicate trouble, weaken the claim.} Low $\hat{J}$ makes corrected estimates unstable; large $|\widehat{\Delta J}|$ makes shared-calibration comparisons unreliable. The appropriate response is not to choose the estimator with the most favorable estimate, but, when possible, to choose designs and estimators based on robustness, and to weaken the strength of the claim being made.
\end{enumerate}

Items 5--7 deserve particular emphasis. The biology case study shows why: with near-zero $\hat{J}$ for one model and a large, well-estimated $\widehat{\Delta J}$, several estimators gave confidently wrong-signed comparisons, and even PPI\texttt{++} produced distorted comparison estimates. The math case shows a second caution: in near-tie comparisons, even modest residual bias can create a false directional claim.

\section{Limitations}

We focus on binary-output LaaJ evaluation, judges that emit binary correctness or preference labels, for single-model accuracy estimation and model comparison. The naive-bias structure described in Section~\ref{sec:naive-and-corrected-est} is broader: ordinal and continuous judge outputs have analogous issues, but the correction and inference framework we study is specific to the binary case. Other settings we do not address include dependent test inputs, adaptive data collection, and broader classes of nonlinear metrics.

Our empirical results are finite-sample and benchmark-specific. The simulations isolate a small number of mechanisms, how estimator quality changes with $J$ and $\Delta J$, so that bias and interval-coverage patterns are interpretable, but they do not span every plausible data-generating regime. The MMLU-Pro case study is informative because it contrasts a relatively well-behaved subject with a failure case, but it does not establish a universal pattern for all benchmarks or judges. In particular, the biology results show that diagnostics can reveal when comparison is unreliable, but they do not provide a complete remedy once judge quality is poor.

Confidence intervals in the MMLU-Pro case study should first be interpreted narrowly, as reflecting uncertainty from the calibration/test partition, finite calibration size, and, where applicable, repeated judge runs or estimator-specific calibration subsampling. A broader population-level interpretation of the $\hat\theta$ intervals additionally requires treating the test items as representative samples from a target population, so that item resampling approximates sampling uncertainty beyond these specific questions. For $\hat{J}$ and $\widehat{\Delta J}$, the narrower interpretation is sufficient: the intervals capture how much the estimated calibration quantities vary with finite calibration data, which is the relevant uncertainty for the diagnostic question.

Calibration-based corrections depend on the quality and representativeness of the calibration set. With small calibration sets, corrected estimates and their intervals can become unstable. This instability is often most severe at low $J$, precisely the regime where careful diagnostics matter most. Beyond finite-sample effects, calibration assumptions can fail in ways not modeled here: for shared calibration, the key risk is the cross-model instability captured by $\Delta J$; for model-specific calibration, mismatch between calibration and test-time responses can also introduce bias that the diagnostics we propose would not detect.

\bibliographystyle{plainnat}
\bibliography{references}

\begin{thebibliography}{24}
\providecommand{\natexlab}[1]{#1}
\providecommand{\url}[1]{\texttt{#1}}
\expandafter\ifx\csname urlstyle\endcsname\relax
  \providecommand{\doi}[1]{doi: #1}\else
  \providecommand{\doi}{doi: \begingroup \urlstyle{rm}\Url}\fi

\bibitem[Angelopoulos et~al.(2023)Angelopoulos, Bates, Fannjiang, Jordan, and
  Zrnic]{angelopoulos2023ppi}
Anastasios~N. Angelopoulos, Stephen Bates, Clara Fannjiang, Michael~I. Jordan,
  and Tijana Zrnic.
\newblock Prediction-powered inference.
\newblock \emph{Science}, 382\penalty0 (6671):\penalty0 669--674, 2023.
\newblock \doi{10.1126/science.adi6000}.
\newblock URL \url{https://www.science.org/doi/abs/10.1126/science.adi6000}.

\bibitem[Angelopoulos et~al.(2024)Angelopoulos, Duchi, and
  Zrnic]{angelopoulos2024ppiefficient}
Anastasios~N. Angelopoulos, John~C. Duchi, and Tijana Zrnic.
\newblock Ppi++: Efficient prediction-powered inference, 2024.
\newblock URL \url{https://arxiv.org/abs/2311.01453}.

\bibitem[Chen et~al.(2026)Chen, Lu, Li, Guo, and Li]{chen2026efficient}
Yiqun~T. Chen, Sizhu Lu, Sijia Li, Moran Guo, and Shengyi Li.
\newblock Efficient inference for noisy llm-as-a-judge evaluation, 2026.
\newblock URL \url{https://arxiv.org/abs/2601.05420}.

\bibitem[Chiang et~al.(2024)Chiang, Zheng, Sheng, Angelopoulos, Li, Li, Zhu,
  Zhang, Jordan, Gonzalez, and Stoica]{chiang2024chatbotarenaopenplatform}
Wei-Lin Chiang, Lianmin Zheng, Ying Sheng, Anastasios~Nikolas Angelopoulos,
  Tianle Li, Dacheng Li, Banghua Zhu, Hao Zhang, Michael Jordan, Joseph~E.
  Gonzalez, and Ion Stoica.
\newblock Chatbot arena: An open platform for evaluating {LLM}s by human
  preference.
\newblock In Ruslan Salakhutdinov, Zico Kolter, Katherine Heller, Adrian
  Weller, Nuria Oliver, Jonathan Scarlett, and Felix Berkenkamp, editors,
  \emph{Proceedings of the 41st International Conference on Machine Learning},
  volume 235 of \emph{Proceedings of Machine Learning Research}, pages
  8359--8388. PMLR, jul 2024.
\newblock URL \url{https://proceedings.mlr.press/v235/chiang24b.html}.

\bibitem[Collot et~al.(2026)Collot, Fraser, Zhao, Shen, Willi, and
  Leontiadis]{collot-etal-2026-balanced}
Stephane Collot, Colin Fraser, Justin Zhao, William~F. Shen, Timon Willi, and
  Ilias Leontiadis.
\newblock Balanced accuracy: The right metric for evaluating {LLM} judges -
  explained through youden{'}s {J} statistic.
\newblock In Yevgen Matusevych, G{\"u}l{\c{s}}en Eryi{\u{g}}it, and Nikolaos
  Aletras, editors, \emph{Proceedings of the 19th Conference of the {E}uropean
  Chapter of the {A}ssociation for {C}omputational {L}inguistics (Volume 5:
  Industry Track)}, pages 927--936, Rabat, Morocco, March 2026. Association for
  Computational Linguistics.
\newblock ISBN 979-8-89176-384-5.
\newblock \doi{10.18653/v1/2026.eacl-industry.69}.
\newblock URL \url{https://aclanthology.org/2026.eacl-industry.69/}.

\bibitem[Dubois et~al.(2023)Dubois, Li, Taori, Zhang, Gulrajani, Ba, Guestrin,
  Liang, and Hashimoto]{dubois2023alpacafarm}
Yann Dubois, Xuechen Li, Rohan Taori, Tianyi Zhang, Ishaan Gulrajani, Jimmy Ba,
  Carlos Guestrin, Percy Liang, and Tatsunori Hashimoto.
\newblock Alpacafarm: A simulation framework for methods that learn from human
  feedback.
\newblock In \emph{Thirty-seventh Conference on Neural Information Processing
  Systems}, 2023.
\newblock URL \url{https://openreview.net/forum?id=4hturzLcKX}.

\bibitem[Dubois et~al.(2024)Dubois, Galambosi, Liang, and
  Hashimoto]{dubois2024length}
Yann Dubois, Bal{\'a}zs Galambosi, Percy Liang, and Tatsunori~B Hashimoto.
\newblock Length-controlled alpacaeval: A simple way to debias automatic
  evaluators.
\newblock \emph{arXiv preprint arXiv:2404.04475}, 2024.

\bibitem[Greiner(2003)]{greiner2003decision}
Matthias Greiner.
\newblock A decision criterion for the application of the rogan-gladen
  estimator in prevalence studies.
\newblock In \emph{Proceedings of the 10th Symposium of the International
  Society for Veterinary Epidemiology and Economics}, pages 240--240, Vina del
  Mar, Chile, 01 2003.

\bibitem[Huang et~al.(2025)Huang, Bu, Zhou, Qu, Liu, Yang, Xu, and
  Zhao]{huang-etal-2025-empirical}
Hui Huang, Xingyuan Bu, Hongli Zhou, Yingqi Qu, Jing Liu, Muyun Yang, Bing Xu,
  and Tiejun Zhao.
\newblock An empirical study of {LLM}-as-a-judge for {LLM} evaluation:
  Fine-tuned judge model is not a general substitute for {GPT}-4.
\newblock In Wanxiang Che, Joyce Nabende, Ekaterina Shutova, and Mohammad~Taher
  Pilehvar, editors, \emph{Findings of the Association for Computational
  Linguistics: ACL 2025}, pages 5880--5895, Vienna, Austria, July 2025.
  Association for Computational Linguistics.
\newblock ISBN 979-8-89176-256-5.
\newblock \doi{10.18653/v1/2025.findings-acl.306}.
\newblock URL \url{https://aclanthology.org/2025.findings-acl.306/}.

\bibitem[Lang and Reiczigel(2014)]{lang2014confidence}
Zsolt Lang and Jenő Reiczigel.
\newblock Confidence limits for prevalence of disease adjusted for estimated
  sensitivity and specificity.
\newblock \emph{Preventive Veterinary Medicine}, 113\penalty0 (1):\penalty0
  13–22, 01 2014.
\newblock \doi{10.1016/j.prevetmed.2013.09.015}.

\bibitem[{LangChain}(n.d.)]{langchain-frozen-cali}
{LangChain}.
\newblock How to calibrate llm-as-a-judge with human corrections, n.d.
\newblock URL \url{https://www.langchain.com/articles/llm-as-a-judge}.
\newblock Accessed: 2026-04-29.

\bibitem[Lee et~al.(2026)Lee, Zeng, Jeong, yong Sohn, and
  Lee]{lee2026correctly}
Chungpa Lee, Thomas Zeng, Jongwon Jeong, Jy~yong Sohn, and Kangwook Lee.
\newblock How to correctly report llm-as-a-judge evaluations, 2026.
\newblock URL \url{https://arxiv.org/abs/2511.21140}.

\bibitem[Liu et~al.(2023)Liu, Iter, Xu, Wang, Xu, and Zhu]{liu-etal-2023-g}
Yang Liu, Dan Iter, Yichong Xu, Shuohang Wang, Ruochen Xu, and Chenguang Zhu.
\newblock {G}-eval: {NLG} evaluation using gpt-4 with better human alignment.
\newblock In Houda Bouamor, Juan Pino, and Kalika Bali, editors,
  \emph{Proceedings of the 2023 Conference on Empirical Methods in Natural
  Language Processing}, pages 2511--2522, Singapore, December 2023. Association
  for Computational Linguistics.
\newblock \doi{10.18653/v1/2023.emnlp-main.153}.
\newblock URL \url{https://aclanthology.org/2023.emnlp-main.153/}.

\bibitem[{Microsoft Azure AI Foundry Blog}(2026)]{microsoft-trusted-judges}
{Microsoft Azure AI Foundry Blog}.
\newblock Evaluating ai agents: Can llm-as-a-judge evaluators be trusted?,
  January 2026.
\newblock URL
  \url{https://techcommunity.microsoft.com/blog/azure-ai-foundry-blog/evaluating-ai-agents-can-llm-as-a-judge-evaluators-be-trusted/4480110}.
\newblock Accessed: 2026-04-29.

\bibitem[Miller(2024)]{miller2024adding}
Evan Miller.
\newblock Adding error bars to evals: A statistical approach to language model
  evaluations, 2024.
\newblock URL \url{https://arxiv.org/abs/2411.00640}.

\bibitem[Peirce(1884)]{peirce1884numerical}
C.~S. Peirce.
\newblock The numerical measure of the success of predictions.
\newblock \emph{Science}, ns-4\penalty0 (93):\penalty0 453--454, 1884.
\newblock \doi{10.1126/science.ns-4.93.453-a}.

\bibitem[Rogan and Gladen(1978)]{rogan1978estimating}
Walter~J. Rogan and Beth Gladen.
\newblock Estimating prevalence from the results of a screening test.
\newblock \emph{American Journal of Epidemiology}, 107\penalty0 (1):\penalty0
  71--76, 1978.
\newblock \doi{10.1093/oxfordjournals.aje.a112510}.

\bibitem[Tan et~al.(2025)Tan, Zhuang, Montgomery, Tang, Cuadron, Wang, Popa,
  and Stoica]{tan2025judgebench}
Sijun Tan, Siyuan Zhuang, Kyle Montgomery, William~Yuan Tang, Alejandro
  Cuadron, Chenguang Wang, Raluca Popa, and Ion Stoica.
\newblock Judgebench: A benchmark for evaluating {LLM}-based judges.
\newblock In \emph{The Thirteenth International Conference on Learning
  Representations}, 2025.
\newblock URL \url{https://openreview.net/forum?id=G0dksFayVq}.

\bibitem[Wang et~al.(2024{\natexlab{a}})Wang, Li, Chen, Cai, Zhu, Lin, Cao,
  Kong, Liu, Liu, and Sui]{wang-etal-2024-large-language-models-fair}
Peiyi Wang, Lei Li, Liang Chen, Zefan Cai, Dawei Zhu, Binghuai Lin, Yunbo Cao,
  Lingpeng Kong, Qi~Liu, Tianyu Liu, and Zhifang Sui.
\newblock Large language models are not fair evaluators.
\newblock In Lun-Wei Ku, Andre Martins, and Vivek Srikumar, editors,
  \emph{Proceedings of the 62nd Annual Meeting of the Association for
  Computational Linguistics (Volume 1: Long Papers)}, pages 9440--9450,
  Bangkok, Thailand, August 2024{\natexlab{a}}. Association for Computational
  Linguistics.
\newblock \doi{10.18653/v1/2024.acl-long.511}.
\newblock URL \url{https://aclanthology.org/2024.acl-long.511/}.

\bibitem[Wang et~al.(2024{\natexlab{b}})Wang, Yu, Yao, Zeng, Yang, Wang, Chen,
  Jiang, Xie, Wang, Xie, Ye, Zhang, and Zhang]{wang2024pandalm}
Yidong Wang, Zhuohao Yu, Wenjin Yao, Zhengran Zeng, Linyi Yang, Cunxiang Wang,
  Hao Chen, Chaoya Jiang, Rui Xie, Jindong Wang, Xing Xie, Wei Ye, Shikun
  Zhang, and Yue Zhang.
\newblock Panda{LM}: An automatic evaluation benchmark for {LLM} instruction
  tuning optimization.
\newblock In \emph{The Twelfth International Conference on Learning
  Representations}, 2024{\natexlab{b}}.
\newblock URL \url{https://openreview.net/forum?id=5Nn2BLV7SB}.

\bibitem[Wang et~al.(2024{\natexlab{c}})Wang, Ma, Zhang, Ni, Chandra, Guo, Ren,
  Arulraj, He, Jiang, Li, Ku, Wang, Zhuang, Fan, Yue, and Chen]{wang2024mmlu}
Yubo Wang, Xueguang Ma, Ge~Zhang, Yuansheng Ni, Abhranil Chandra, Shiguang Guo,
  Weiming Ren, Aaran Arulraj, Xuan He, Ziyan Jiang, Tianle Li, Max Ku, Kai
  Wang, Alex Zhuang, Rongqi Fan, Xiang Yue, and Wenhu Chen.
\newblock Mmlu-pro: A more robust and challenging multi-task language
  understanding benchmark.
\newblock In A.~Globerson, L.~Mackey, D.~Belgrave, A.~Fan, U.~Paquet,
  J.~Tomczak, and C.~Zhang, editors, \emph{Advances in Neural Information
  Processing Systems}, volume~37, pages 95266--95290. Curran Associates, Inc.,
  2024{\natexlab{c}}.
\newblock \doi{10.52202/079017-3018}.
\newblock URL
  \url{https://proceedings.neurips.cc/paper_files/paper/2024/file/ad236edc564f3e3156e1b2feafb99a24-Paper-Datasets_and_Benchmarks_Track.pdf}.

\bibitem[Xie et~al.(2025)Xie, Li, Yu, Zhang, Zhang, and Yang]{xie2025an}
Qiujie Xie, Qingqiu Li, Zhuohao Yu, Yuejie Zhang, Yue Zhang, and Linyi Yang.
\newblock An empirical analysis of uncertainty in large language model
  evaluations.
\newblock In \emph{The Thirteenth International Conference on Learning
  Representations}, 2025.
\newblock URL \url{https://openreview.net/forum?id=J4xLuCt2kg}.

\bibitem[Youden(1950)]{youden1950index}
W.~J. Youden.
\newblock Index for rating diagnostic tests.
\newblock \emph{Cancer}, 3\penalty0 (1):\penalty0 32--35, 1950.
\newblock \doi{10.1002/1097-0142(1950)3:1<32::AID-CNCR2820030106>3.0.CO;2-3}.

\bibitem[Zheng et~al.(2023)Zheng, Chiang, Sheng, Zhuang, Wu, Zhuang, Lin, Li,
  Li, Xing, Zhang, Gonzalez, and Stoica]{zheng-laaj}
Lianmin Zheng, Wei-Lin Chiang, Ying Sheng, Siyuan Zhuang, Zhanghao Wu, Yonghao
  Zhuang, Zi~Lin, Zhuohan Li, Dacheng Li, Eric Xing, Hao Zhang, Joseph~E
  Gonzalez, and Ion Stoica.
\newblock Judging llm-as-a-judge with mt-bench and chatbot arena.
\newblock In A.~Oh, T.~Naumann, A.~Globerson, K.~Saenko, M.~Hardt, and
  S.~Levine, editors, \emph{Advances in Neural Information Processing Systems},
  volume~36, pages 46595--46623. Curran Associates, Inc., 2023.
\newblock URL
  \url{https://proceedings.neurips.cc/paper_files/paper/2023/file/91f18a1287b398d378ef22505bf41832-Paper-Datasets_and_Benchmarks.pdf}.

\end{thebibliography}

\appendix

\section{Calibration structure, RG vs PPI\texttt{++}} \label{appendix:rg-vs-ppi-calibration}

\subsection{Judge-centric vs model-specific} \label{appendix:cali-estimator-focus}

A key structural difference between RG and PPI\texttt{++} is the role of the calibration set. In RG, calibration is \textbf{judge-centric}: the calibration set consists of examples with known ground-truth labels, and the judge is run on those examples to estimate its sensitivity $q_1$ and specificity $q_0$. These quantities characterize the judge independently of the model being evaluated. Under the model-independence assumption, $\hat{q}_0$ and $\hat{q}_1$ can be estimated once and reused for any model: adding a new model requires only running the judge on its outputs, with no additional human labeling. In contrast, PPI\texttt{++} is \textbf{model-specific}: the calibration set pairs true labels with the outputs of the specific model under test, and the resulting correction is tied to that model. Evaluating a new model generally requires a new calibration set, which can be a substantial ongoing cost in workflows where models change frequently.

Both estimators can accommodate replacing the judge without new human labeling, provided the new judge is rerun on the existing labeled calibration set. The model-specificity constraint still applies: for PPI\texttt{++}, the calibration set must come from the model being evaluated, while for RG it is enough that $q_0$ and $q_1$ remain stable between calibration and test.

When model-specific calibration is used, the variance-minimizing $\lambda^* \approx \theta(1-\theta)J / [p(1-p)]$. As $J \to 0$, $\lambda^* \to 0$: PPI\texttt{++} gives the correction term little weight and falls back on the labeled calibration mean $\bar{Y}_\mathrm{cal}$. This is graceful degradation, as precision is limited by calibration-set size, but there is no amplification of noise. The contrast with shared-calibration PPI\texttt{++} (and with RG) is instructive: the subsection below shows that shared calibration forces $\lambda = 1/J$, which explodes as $J \to 0$ and is the source of the instability described in the main text.

\subsection{PPI\texttt{++} with shared calibration reduces to the RG estimator} \label{appendix:ppi-shared-cali}

The main text argues that PPI\texttt{++} requires per-model calibration while the RG estimator does not. A natural question is whether PPI\texttt{++} could simply adopt the same model-independence assumption as RG, reusing a single calibration set across models. The answer is yes, but doing so turns PPI\texttt{++} into the RG estimator. The role of $\lambda$ must also change: instead of minimizing variance, it must be chosen to debias the estimator.

\paragraph{Setup.} Suppose we have calibration data from model $C$ (ground-truth labels paired with judge scores), and we want to estimate $\theta_A$, the accuracy of a different model $A$, using only judge scores on model $A$'s test outputs. Under model-independence, the judge's sensitivity $q_1$ and specificity $q_0$ are the same for both models. The shared-calibration PPI\texttt{++} estimator takes the form
$$
\hat\theta_A = \bar Y_C + \lambda(\hat p_A - \hat p_C),
$$
where $\bar Y_C$ is the mean ground-truth accuracy in the calibration set, $\hat p_A$ is the mean judge score on model $A$'s test outputs, and $\hat p_C$ is the mean judge score on the calibration data (model $C$'s outputs).

\paragraph{Unbiasedness condition.} For the estimator to be unbiased for $\theta_A$ we need $\mathbb{E}[\hat\theta_A] = \theta_A$, i.e.,
$$
\theta_C + \lambda(p_A - p_C) = \theta_A.
$$
Since $p = \theta J + (1 - q_0)$, we have $p_A - p_C = J(\theta_A - \theta_C)$, and so the required tuning parameter is
$$
\lambda = \frac{1}{J},
$$
which depends only on the judge, not on either model. This is the same judge-only quantity that appears in the RG estimator.

\paragraph{Algebraic equivalence.} Plugging in $\lambda = 1/\hat J$, where $\hat J = \hat q_0 + \hat q_1 - 1$, gives
$$
\hat\theta_A = \bar Y_C + \frac{1}{\hat J}(\hat p_A - \hat p_C).
$$
Because $\bar Y_C$, $\hat p_C$, $\hat q_0$, and $\hat q_1$ are all computed from the same calibration sample, they satisfy the empirical identity
$$
\hat p_C = \bar Y_C \hat q_1 + (1-\bar Y_C)(1-\hat q_0),
$$
which implies
$$
\bar Y_C = \frac{\hat p_C + \hat q_0 - 1}{\hat J}.
$$
Substituting this into the estimator gives
\begin{align*}
	\hat\theta_A
	&= \frac{\hat p_C + \hat q_0 - 1}{\hat J} + \frac{\hat p_A - \hat p_C}{\hat J} \\
	&= \frac{\hat p_A + \hat q_0 - 1}{\hat J} \\
	&= \hat\theta_{\mathrm{RG},A}.
\end{align*}
Thus, under this shared-calibration setup, cross-model PPI\texttt{++} reduces exactly to the RG estimator.

\paragraph{Implication.} Under the model-independence assumption, the only signal shared across models is the judge's error rates $(q_0, q_1)$. RG uses that signal directly, and shared-calibration PPI\texttt{++} collapses to the same estimator. The efficiency advantage of PPI\texttt{++} over RG arises only when it uses model-specific calibration data, but that also incurs a per-model labeling cost that RG avoids.

\paragraph{The $1/J$ factor is inescapable under shared calibration.}
Equation~\eqref{eq:measurement-model} can be rewritten as $p = \theta J + (1 - q_0)$, so under shared calibration $\Delta p = \Delta\theta \cdot J$, giving $\Delta\theta = \Delta p / J$. The derivation above shows that $\lambda = 1/J$ is the value forced on any estimator that correctly recovers $\Delta\theta$ from $\Delta p$ while sharing calibration across models. This is not a property of RG or PPI\texttt{++} specifically; it is a consequence of the measurement model. Low $J$ attenuates the signal in $\Delta p$, and recovering $\Delta\theta$ requires undoing that attenuation regardless of the estimator chosen.

\section{Analytical variance for the RG comparison estimator} \label{appendix:rg-variance-estimator}

This appendix derives the variance for the shared-calibration RG comparison estimator, showing how $1/J^2$ scaling enters and why pairing test-set items matters.

\paragraph{Shared-calibration simplification.}
When both models share calibration estimates $(\hat{q}_0, \hat{q}_1)$, the $\hat{q}_0$ terms cancel in the difference:
$$
\widehat{\Delta\theta}_\mathrm{RG}
    = \frac{\hat{p}_B + \hat{q}_0 - 1}{\hat{J}} - \frac{\hat{p}_A + \hat{q}_0 - 1}{\hat{J}}
    = \frac{\hat{p}_B - \hat{p}_A}{\hat{J}},
$$
where $\hat{J} = \hat{q}_0 + \hat{q}_1 - 1$. Only uncertainty in $\hat{J}$ survives from the calibration side. Applying the delta method:
$$
\mathrm{SE}^2 \approx
    \underbrace{
        \frac{\widehat{\mathrm{Var}}(\hat{Z}_{B,i} - \hat{Z}_{A,i})}{N \cdot \hat{J}^2}
    }_{\text{test}}
    +
    \underbrace{
        \frac{
            (\hat{p}_B - \hat{p}_A)^2
            \left(\dfrac{\hat{q}_0(1-\hat{q}_0)}{m_0} + \dfrac{\hat{q}_1(1-\hat{q}_1)}{m_1}\right)
        }{\hat{J}^4}
    }_{\text{calibration}}.
$$
The test term scales as $1/\hat J^2$ and the calibration term as $1/\hat J^4$, so small $\hat J$ sharply inflates interval width.

\paragraph{Pairing.}
Estimating the test term from paired per-input differences $\hat{Z}_{B,i} - \hat{Z}_{A,i}$ is preferable to treating the two models' scores as independent. The paired estimate captures model correlation directly and avoids a component-wise expansion that can go negative when models are highly correlated. The bootstrap preserves this pairing automatically by resampling both models on the same input indices in each replicate.

\section{Estimation of bias in $\theta_\mathrm{RG}$ when sharing calibration} \label{appendix:rg-shared-cali-bias}

When we estimate $\theta$ with model A's own calibration labels, we get
$$
\hat\theta_{\mathrm{RG,A}} = \frac{\hat p_A + \hat q_{0,A} - 1}{\hat q_{0,A} + \hat q_{1,A} - 1},
$$

but when we instead use $\hat q_{0,B}$ and $\hat q_{1,B}$ estimated from model B's calibration labels we get

$$
\hat\theta_{\mathrm{RG,A}}^* = \frac{\hat p_A + \hat q_{0,B} - 1}{\hat q_{0,B} + \hat q_{1,B} - 1}.
$$

Since $\hat\theta_{\mathrm{RG,A}}$ is asymptotically unbiased for $\theta_A$, the asymptotic bias of $\hat\theta_{\mathrm{RG,A}}^*$, the limit of $\hat\theta_{\mathrm{RG,A}}^* - \hat\theta_{\mathrm{RG,A}}$ as calibration-set size grows, is

$$
\begin{aligned}
		\text{bias} &= \frac{p_A + q_{0,B} - 1}{J_B} - \frac{p_A + q_{0,A} - 1}{J_A} \\
		&\qquad \text{with } \Delta J = J_A - J_B \text{ and } \Delta q_0 = q_{0,A} - q_{0,B} \\
		&= \frac{\theta_A \cdot \Delta J - \Delta q_0}{J_B} \\
		&= \frac{\theta_A \cdot \Delta q_1}{J_B} - \frac{(1 - \theta_A) \Delta q_0}{J_B}
\end{aligned}
$$

Thus calibration mismatch between models is amplified by $1/J$ into comparison bias. When $J$ is small, this amplification is severe. When $J$ differs between models, the amplification creates directional comparison bias even if the difference is not statistically significant. In the middle line, the $\theta_A \cdot \Delta J$ term means that models with higher accuracy are more sensitive to calibration mismatch. The last line shows, since $0.5 < \theta_A \leq 1.0$, that $\Delta q_1$ contributes more than $\Delta q_0$, and the two terms partially cancel when $\Delta q_1$ and $\Delta q_0$ have the same sign.

Note that $\theta_B$ will continue to be unbiased, so all of the bias occurs in $\hat\theta_{\mathrm{RG,A}}^*$, which gets carried over to $\widehat{\Delta\theta}$.

Using the numbers from Mistral Large on the math questions
$$
\begin{aligned}
\Delta J &= -0.119 \\
\Delta q_0 &= -0.060 \\
J_\text{Qwen} &= 0.519 \\
\theta_{\text{Gemma\,3}} &\approx 0.738 \\
\text{bias} & \approx (0.738 \times (-0.119) - (-0.060)) / 0.519 \\
& \approx -0.054 \\
\widehat{\theta}_{\text{Gemma\,3}} - \theta_{\text{Gemma\,3}} &\approx 0.665 - 0.738 = -0.073
\end{aligned}
$$

Hence this bias can explain the majority of the error in the estimate of $\theta_{\text{Gemma\,3}}$, with the remainder ($-0.019$) being finite-sample noise in the calibration estimates themselves.

\section{Additional law and engineering subjects} \label{appendix:more-laaj-results}

Law and engineering add two intermediate regimes. Law shows that low $J$ alone can severely distort naive per-model estimates and make RG-based correction unstable, even when $\widehat{\Delta J}$ is not clearly different from zero. Engineering is more mixed: under Mistral Large, naive and shared-calibration RG yield substantially exaggerated comparison estimates, while under Gemma\,4\,31B the comparison estimate is better only because the per-model biases largely cancel. PPI\texttt{++} remains relatively well behaved in both subjects. In each subject, one of the naive comparison estimates is accurate by luck, due to per-model biases that almost entirely cancel.

Table~\ref{tab:judge-J-additional} reports the judge diagnostics. For law, Mistral Large has very low $J$ for both models, while Gemma\,4\,31B is better but still moderate. In both cases the $\widehat{\Delta J}$ intervals include zero, so this subject isolates a low-$J$ failure mode more than a dramatic cross-model-instability story. Engineering shows moderate judge quality overall, with the $\widehat{\Delta J}$ intervals again including zero.

\begin{table}[t]
	\centering
	\resizebox{\textwidth}{!}{
		\begin{tabular}{llcccc}
			\toprule
			Subject & Judge & $\hat{J}_{\mathrm{Gemma\,3}}$ & $\hat{J}_{\mathrm{Qwen\,2.5}}$ & $\widehat{\Delta J}$ & $\widehat{\Delta J}$ plotted
			\\
			\midrule
			\multirow{4}{*}{Law}
			& \multirow{2}{*}{Mistral Large}
			& $0.142$ & $0.184$ & $-0.042$
			& \multirow{2}{*}{\jplot{-0.152}{0.071}{-0.042}} \\
			& & \small$[0.059,\ 0.222]$
			& \small$[0.095,\ 0.269]$
			& \small$[{-0.152},\ 0.071]$ & \\[4pt]
			& \multirow{2}{*}{Gemma\,4\,31B}
			& $0.393$ & $0.486$ & $-0.094$
			& \multirow{2}{*}{\jplot{-0.212}{0.026}{-0.094}} \\
			& & \small$[0.297,\ 0.486]$
			& \small$[0.399,\ 0.573]$
			& \small$[{-0.212},\ 0.026]$ & \\
			\midrule
			\multirow{4}{*}{Engineering}
			& \multirow{2}{*}{Mistral Large}
			& $0.342$ & $0.446$ & $-0.104$
			& \multirow{2}{*}{\jplot{-0.245}{0.038}{-0.104}} \\
			& & \small$[0.237,\ 0.442]$
			& \small$[0.343,\ 0.547]$
			& \small$[{-0.245},\ 0.038]$ & \\[4pt]
			& \multirow{2}{*}{Gemma\,4\,31B}
			& $0.366$ & $0.428$ & $-0.062$
			& \multirow{2}{*}{\jplot{-0.176}{0.047}{-0.062}} \\
			& & \small$[0.270,\ 0.456]$
			& \small$[0.332,\ 0.524]$
			& \small$[{-0.176},\ 0.047]$ & \\
			\bottomrule
		\end{tabular}
	}
	\caption{Per-model Youden's $\hat{J} = \hat{q}_0 + \hat{q}_1 - 1$ and difference
		$\widehat{\Delta J}$ with 95\% bootstrap CIs for additional law and engineering subjects. Right column shows $\widehat{\Delta J}$ on the same scale as Table~\ref{tab:judge-J}; dotted line marks zero.}
	\label{tab:judge-J-additional}
\end{table}

\begin{figure}[t]
	\centering
	\begin{subfigure}[t]{0.500\textwidth}
		\centering
		\includegraphics[width=\linewidth]{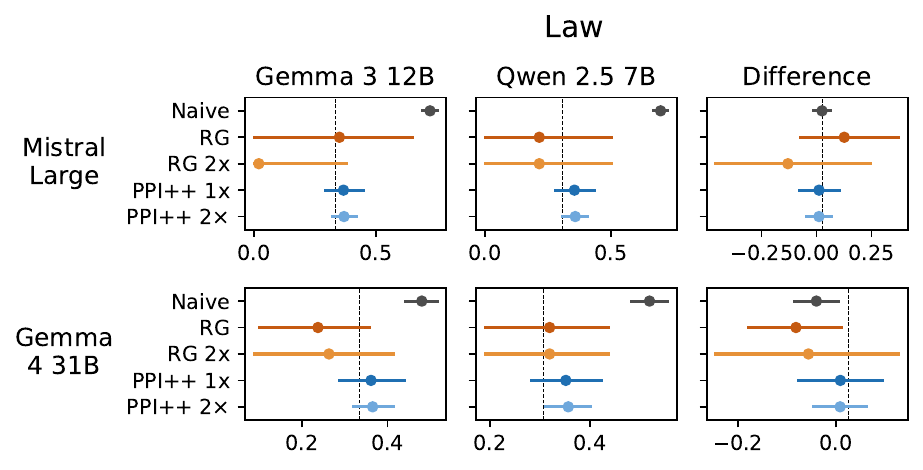}
	\end{subfigure}
	\begin{subfigure}[t]{0.490\textwidth}
		\centering
		\includegraphics[width=\linewidth]{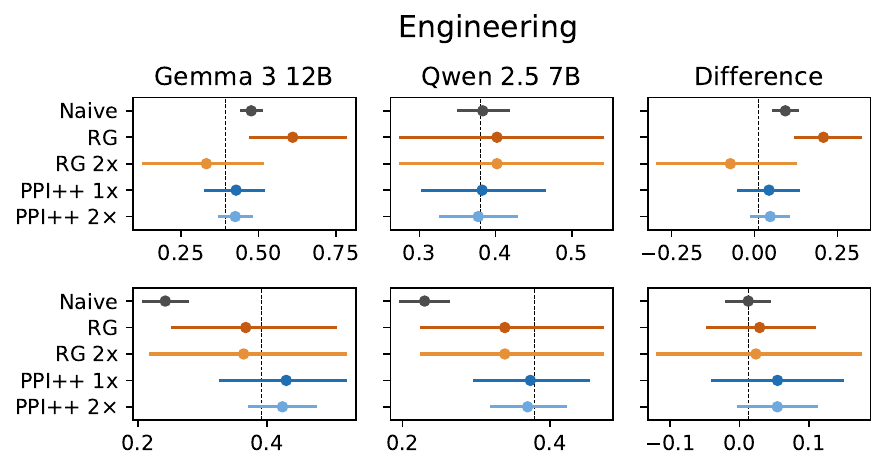}
	\end{subfigure}
	\caption{Bootstrap medians and 95\% confidence intervals for per-model accuracy $\hat\theta$ for the naive estimator and four corrected estimators on the additional law and engineering subjects. The dotted vertical line shows true accuracy (mean on the test set).}
	\label{fig:judge-cis-additional}
\end{figure}

Table~\ref{tab:judge-theta-diffs-additional} reports comparison estimates, and Figure~\ref{fig:judge-cis-additional} shows the corresponding per-model accuracy estimates.

In law, the true comparison is a small positive gap ($+0.026$). The most striking pattern is the per-model bias: under both judges, the naive estimator substantially overestimates both models' accuracies. For Mistral Large, those biases are nearly equal in magnitude and largely cancel in the difference, so the naive comparison contains the truth by coincidence rather than by reliability. For Gemma\,4\,31B, the biases do not cancel as much, and the naive comparison is mildly wrong-signed at the point-estimate level; a reminder that even moderate judge quality does not guarantee a reliable comparison estimate. RG remains relatively unstable, with fairly wide intervals. PPI\texttt{++} again gives the most stable comparison estimates, with medians near the truth and intervals containing it for both judges and both calibration budgets.

In engineering, the true comparison is smaller ($+0.012$), making the subject another near-tie regime. For Mistral Large, the naive estimator overestimates Gemma\,3's performance but is close to the truth for Qwen\,2.5, resulting in a comparison estimate and interval that lie entirely above the true difference. Shared-calibration RG shows the same pattern, with an even more exaggerated estimate. For Gemma\,4\,31B, the naive estimator underestimates both base models by roughly the same amount, resulting in a comparison estimate that is very accurate, but only by luck. This is very similar to the naive estimator result for law with Mistral Large, but with base-model underestimation rather than overestimation. PPI\texttt{++} is again relatively well behaved for engineering.

\begin{table}[t]
	\centering
	\small
	\setlength{\tabcolsep}{4pt}
	\resizebox{\textwidth}{!}{
		\begin{tabular}{ll c ccccc}
			\toprule
			Subject & Judge & $\Delta\theta$
			& $\widehat{\Delta\theta}_{\mathrm{naive}}$
			& $\widehat{\Delta\theta}_{\mathrm{RG}}$
			& $\widehat{\Delta\theta}_{\mathrm{RG,2\times}}$
			& $\widehat{\Delta\theta}_{\mathrm{PPI{+}{+}}}$
			& $\widehat{\Delta\theta}_{\mathrm{PPI{+}{+},2\times}}$ \\
			\midrule
			\multirow{4}{*}{Law}
			& \multirow{2}{*}{Mistral Large} & \multirow{2}{*}{$+$0.026}
			& $+$0.025 & $+$0.126 & $-$0.130 & $+$0.011 & $+$0.011 \\
			& & & \footnotesize$[-0.014,\;+0.064]$
			& \footnotesize$[-0.071,\;+0.374]$
			& \footnotesize$[-0.457,\;+0.245]$
			& \footnotesize$[-0.077,\;+0.103]$
			& \footnotesize$[-0.045,\;+0.068]$ \\[4pt]
			& \multirow{2}{*}{Gemma\,4\,31B} & \multirow{2}{*}{$+$0.026}
			& $-$0.040 & $-$0.081 & $-$0.056 & $+$0.010 & $+$0.009 \\
			& & & \footnotesize$[-0.083,\;+0.005]$
			& \footnotesize$[-0.177,\;+0.011]$
			& \footnotesize$[-0.245,\;+0.129]$
			& \footnotesize$[-0.076,\;+0.096]$
			& \footnotesize$[-0.045,\;+0.064]$ \\
			\midrule
			\multirow{4}{*}{Engineering}
			& \multirow{2}{*}{Mistral Large} & \multirow{2}{*}{$+$0.012}
			& $+$0.093 & $+$0.208 & $-$0.073 & $+$0.044 & $+$0.047 \\
			& & & \footnotesize$[+0.056,\;+0.128]$
			& \footnotesize$[+0.122,\;+0.319]$
			& \footnotesize$[-0.291,\;+0.122]$
			& \footnotesize$[-0.049,\;+0.133]$
			& \footnotesize$[-0.010,\;+0.102]$ \\[4pt]
			& \multirow{2}{*}{Gemma\,4\,31B} & \multirow{2}{*}{$+$0.012}
			& $+$0.012 & $+$0.029 & $+$0.023 & $+$0.055 & $+$0.054 \\
			& & & \footnotesize$[-0.019,\;+0.043]$
			& \footnotesize$[-0.046,\;+0.108]$
			& \footnotesize$[-0.118,\;+0.174]$
			& \footnotesize$[-0.039,\;+0.148]$
			& \footnotesize$[-0.001,\;+0.110]$ \\
			\bottomrule
		\end{tabular}
	}
	\caption{Gemma\,3 minus Qwen\,2.5 accuracy on additional law and engineering subjects. $\Delta\theta$ is the true difference. Estimates are bootstrap medians with 95\% bootstrap CIs. RG uses shared calibration; RG\,$2\times$ uses model-specific calibration with the RG formula; PPI\texttt{++} variants use model-specific calibration.}
	\label{tab:judge-theta-diffs-additional}
\end{table}

Taken together, these additional subjects sharpen a few practical lessons. First, low judge quality alone can be enough to make naive and RG-based estimates unreliable; a non-alarming $\widehat{\Delta J}$ estimate is not enough by itself to make the comparison trustworthy. Second, near-tie comparisons can sometimes look acceptable because per-model biases cancel, but such cancellation is fragile and can mask underlying estimation problems if the comparison estimate is viewed in isolation. Finally, as with math and biology, PPI\texttt{++} with model-specific calibration was the most consistently well-behaved estimator among those we study.

\section{MMLU-Pro details for LLM-as-a-Judge} \label{appendix:mmlu-pro-laaj}

\subsection{System prompt and structured output}

The system prompt used for all judge evaluations was:
\begin{quote}
\small
You are evaluating a student's reasoning on a multiple-choice question. You will see the question and the student's complete reasoning, but NOT their final answer. Assess whether the student's reasoning directly implies the correct answer. First give your own reasoning for the problem in \texttt{reasoning}, then give your assessment of the student's reasoning: ``Yes'' if the student's reasoning is accurate and sufficient to arrive at the correct answer, ``No'' if it is not.
\end{quote}

Structured outputs were required, using the following schema:
\begin{lstlisting}[language=Python]
{
    "type": "json_schema",
    "json_schema": {
        "name": "JudgeOutput",
        "strict": True,
        "schema": {
            "type": "object",
            "properties": {
                "reasoning": {"type": "string"},
                "prediction": {"type": "string", "enum": ["Yes", "No"]},
            },
            "required": ["reasoning", "prediction"],
            "additionalProperties": False,
        },
    },
}
\end{lstlisting}

\subsection{Base model selection and answer extraction}

Qwen~2.5 7B and Gemma~3 12B were selected as base models because of their similar performance across MMLU-Pro subjects studied here. Responses were generated using a CoT prompt similar to \citet{wang2024mmlu}. We collected $k=5$ outputs per question for each model; not all runs produced correctly-formatted answers, so the first correctly-formatted output per question was selected as the single input used here. Questions for which neither Qwen\,2.5\,7B nor Gemma\,3\,12B produced any correctly-formatted response across all five runs were dropped entirely. Table~\ref{tab:base-model-questions} shows question counts before and after this filtering.

\begin{table}[h]
	\centering
	\small
	\caption{Questions per subject before and after filtering for at least one correctly formatted response across $k=5$ runs.}
	\label{tab:base-model-questions}
	\begin{tabular}{lrrrr}
		\toprule
		& Math & Biology & Law & Engineering \\
		\midrule
		In benchmark           & 1351 & 717 & 1101 & 969 \\
		At least one correctly formatted response & 1290 & 717 & 1101 & 963 \\
		\bottomrule
	\end{tabular}
\end{table}

Each base model's full response was given to the judge models, except that the final answer letter was removed. The final answer is identified as the last match of the first regular expression below; if no match is found, the last match of the second is used. If neither matches, the response is considered incorrectly formatted and was excluded.
\begin{lstlisting}[language=Python]
ANSWER_RE_1 = re.compile(
    r"(?:T|t)he [a-z ]*answer is[a-z ]*[:\s]*\(?([A-Ja-j])\)?"
)
ANSWER_RE_2 = re.compile(r"(?:A|a)nswer:\s+\(?([A-Ja-j])\)?")
\end{lstlisting}

This procedure does not guard against a base model's answer being obvious from the remaining text, for example if it was repeated earlier in the response. For present purposes the procedure is sufficient: $J$ ranges from 0.09 to 0.67 across judges and subjects (Table~\ref{tab:judge-J}), indicating moderate rather than near-perfect discriminative ability.

\subsection{Calibration and test sets}

We randomly selected one-third of the questions as a calibration set and two-thirds as a test set. The four estimators use the calibration set as follows.

\textbf{RG} uses the entire calibration set with true labels from Qwen\,2.5\,7B only, applying the resulting calibration quantities to both models. This simulates Qwen\,2.5\,7B as the incumbent model with pre-existing labels, and Gemma\,3\,12B as a challenger with no additional labeling budget. This is our base estimator.

\textbf{RG}\,$\mathbf{2\times}$ uses the full calibration set with labels from both models, creating separate per-model calibration estimates. This simulates twice the labeling budget of base RG, and is comparable to PPI\texttt{++}\,$2\times$.

\textbf{PPI\texttt{++}}\,$\mathbf{1\times}$ simulates the same total labeling budget as base RG: labels from both Qwen\,2.5\,7B and Gemma\,3\,12B are used, but only on a randomly selected half of the calibration set. The random selection introduces additional variability, which is handled in the bootstrap procedure below.

\textbf{PPI\texttt{++}}\,$\mathbf{2\times}$ uses labels from both models on the entire calibration set, matching the labeling budget of RG\,$2\times$.

For the judge models, $k=1$ is used for all calibration sets. On the test set, Mistral Large uses $k=5$ runs on the same base model inputs; Gemma\,4\,31B uses $k=1$, as it was observed to produce highly consistent outputs in preliminary testing. Nondeterminism in Mistral Large aligns with findings elsewhere that LLM-as-a-Judge can be inconsistent across runs (\citep{microsoft-trusted-judges}).

\subsection{Bootstrap procedure}

For all estimators, calibration and test sets were separately sampled (reflecting a fixed split) across 10,000 bootstrap iterations. The calibration and test sets both represent sources of variance, so both needed to be sampled. The confidence intervals are the $\alpha / 2$ and $1 - \alpha / 2$ quantiles.

PPI\texttt{++}\,$1\times$ has an extra source of variance due to randomly sampling half the calibration set. To average out this extra variance, bootstrapping was repeated 20 times, with 20 separate selections of the calibration subset. The resulting bootstrap iterations were pooled to create estimates. This extra randomness also makes a plug-in point estimate suboptimal. We instead use the bootstrap median, and apply it to all $\theta$ estimates for within-table consistency.

PPI\texttt{++}'s $\lambda$ estimation presents a challenge with $k > 1$: passing $k \cdot n_\text{test}$ run-level rows would set $N = k \cdot n_\text{test}$ in the $\lambda$ formula, underscaling $\mathrm{Var}(\hat{Y}_\text{test}) / N$ by a factor of $k$ and producing an upward-biased $\lambda$. Hence, Mistral's $k=5$ runs in the test set were collapsed to a per-question mean accuracy, and then IID bootstrap was used.

\subsection{Full results}

Table~\ref{tab:judge-q0q1} gives per-model $\hat{q}_0$ and $\hat{q}_1$ with 95\% bootstrap CIs for each judge, base model, and subject. Table~\ref{tab:judge-theta-estimates} gives complete numerical results for per-model accuracy estimates and comparison differences.

\begin{table}[p]
	\centering
		\begin{tabular}{lllcc}
			\toprule
			Subject & Judge & Base model & $\hat{q}_0$ & $\hat{q}_1$ \\
			\midrule
			
			\multirow{12}{*}{Math}
			& \multirow{6}{*}{Mistral Large}
			& \multirow{2}{*}{Gemma\,3\,12B}
			& $0.509$ & $0.887$ \\
			& & & \small$[0.415,\ 0.602]$ & \small$[0.852,\ 0.920]$ \\[2pt]
			& & \multirow{2}{*}{Qwen\,2.5\,7B}
			& $0.569$ & $0.947$ \\
			& & & \small$[0.473,\ 0.661]$ & \small$[0.921,\ 0.970]$ \\[2pt]
			& & \multirow{2}{*}{$\Delta$}
			& $-0.060$ & $-0.060$ \\
			& & & \small$[{-0.186},\ 0.070]$ & \small$[{-0.099},\ {-0.021}]$ \\
			\cmidrule(l){2-5}
			
			& \multirow{6}{*}{Gemma\,4\,31B}
			& \multirow{2}{*}{Gemma\,3\,12B}
			& $0.764$ & $0.878$ \\
			& & & \small$[0.680,\ 0.843]$ & \small$[0.841,\ 0.913]$ \\[2pt]
			& & \multirow{2}{*}{Qwen\,2.5\,7B}
			& $0.752$ & $0.919$ \\
			& & & \small$[0.669,\ 0.833]$ & \small$[0.888,\ 0.948]$ \\[2pt]
			& & \multirow{2}{*}{$\Delta$}
			& $+0.011$ & $-0.041$ \\
			& & & \small$[{-0.087},\ 0.108]$ & \small$[{-0.087},\ 0.005]$ \\
			\midrule
			
			\multirow{12}{*}{Biology}
			& \multirow{6}{*}{Mistral Large}
			& \multirow{2}{*}{Gemma\,3\,12B}
			& $0.348$ & $0.746$ \\
			& & & \small$[0.211,\ 0.488]$ & \small$[0.684,\ 0.806]$ \\[2pt]
			& & \multirow{2}{*}{Qwen\,2.5\,7B}
			& $0.395$ & $0.988$ \\
			& & & \small$[0.288,\ 0.507]$ & \small$[0.969,\ 1.000]$ \\[2pt]
			& & \multirow{2}{*}{$\Delta$}
			& $-0.047$ & $-0.242$ \\
			& & & \small$[{-0.201},\ 0.103]$ & \small$[{-0.307},\ {-0.179}]$ \\
			\cmidrule(l){2-5}
			
			& \multirow{6}{*}{Gemma\,4\,31B}
			& \multirow{2}{*}{Gemma\,3\,12B}
			& $0.435$ & $0.746$ \\
			& & & \small$[0.293,\ 0.581]$ & \small$[0.684,\ 0.806]$ \\[2pt]
			& & \multirow{2}{*}{Qwen\,2.5\,7B}
			& $0.553$ & $0.975$ \\
			& & & \small$[0.442,\ 0.662]$ & \small$[0.949,\ 0.994]$ \\[2pt]
			& & \multirow{2}{*}{$\Delta$}
			& $-0.118$ & $-0.229$ \\
			& & & \small$[{-0.262},\ 0.025]$ & \small$[{-0.293},\ {-0.166}]$ \\
			\midrule
			
			\multirow{12}{*}{Law}
			& \multirow{6}{*}{Mistral Large}
			& \multirow{2}{*}{Gemma\,3\,12B}
			& $0.279$ & $0.862$ \\
			& & & \small$[0.222,\ 0.339]$ & \small$[0.801,\ 0.917]$ \\[2pt]
			& & \multirow{2}{*}{Qwen\,2.5\,7B}
			& $0.342$ & $0.842$ \\
			& & & \small$[0.282,\ 0.401]$ & \small$[0.777,\ 0.903]$ \\[2pt]
			& & \multirow{2}{*}{$\Delta$}
			& $-0.062$ & $+0.020$ \\
			& & & \small$[{-0.139},\ 0.014]$ & \small$[{-0.065},\ 0.102]$ \\
			\cmidrule(l){2-5}
			
			& \multirow{6}{*}{Gemma\,4\,31B}
			& \multirow{2}{*}{Gemma\,3\,12B}
			& $0.624$ & $0.768$ \\
			& & & \small$[0.560,\ 0.687]$ & \small$[0.697,\ 0.837]$ \\[2pt]
			& & \multirow{2}{*}{Qwen\,2.5\,7B}
			& $0.637$ & $0.850$ \\
			& & & \small$[0.574,\ 0.697]$ & \small$[0.787,\ 0.910]$ \\[2pt]
			& & \multirow{2}{*}{$\Delta$}
			& $-0.012$ & $-0.082$ \\
			& & & \small$[{-0.094},\ 0.068]$ & \small$[{-0.170},\ 0.006]$ \\
			\midrule
			
			\multirow{12}{*}{Engineering}
			& \multirow{6}{*}{Mistral Large}
			& \multirow{2}{*}{Gemma\,3\,12B}
			& $0.637$ & $0.705$ \\
			& & & \small$[0.566,\ 0.707]$ & \small$[0.627,\ 0.779]$ \\[2pt]
			& & \multirow{2}{*}{Qwen\,2.5\,7B}
			& $0.796$ & $0.650$ \\
			& & & \small$[0.740,\ 0.851]$ & \small$[0.562,\ 0.733]$ \\[2pt]
			& & \multirow{2}{*}{$\Delta$}
			& $-0.159$ & $+0.055$ \\
			& & & \small$[{-0.245},\ {-0.073}]$ & \small$[{-0.052},\ 0.162]$ \\
			\cmidrule(l){2-5}
			
			& \multirow{6}{*}{Gemma\,4\,31B}
			& \multirow{2}{*}{Gemma\,3\,12B}
			& $0.891$ & $0.475$ \\
			& & & \small$[0.844,\ 0.933]$ & \small$[0.392,\ 0.558]$ \\[2pt]
			& & \multirow{2}{*}{Qwen\,2.5\,7B}
			& $0.915$ & $0.512$ \\
			& & & \small$[0.876,\ 0.951]$ & \small$[0.423,\ 0.600]$ \\[2pt]
			& & \multirow{2}{*}{$\Delta$}
			& $-0.025$ & $-0.038$ \\
			& & & \small$[{-0.078},\ 0.028]$ & \small$[{-0.131},\ 0.051]$ \\
			\bottomrule
		\end{tabular}
	\caption{Per-model specificity $\hat{q}_0$ and sensitivity $\hat{q}_1$ with 95\% bootstrap CIs, by subject, judge, and base model. $\Delta = \hat{q}_{\mathrm{Gemma\,3}} - \hat{q}_{\mathrm{Qwen\,2.5}}$.}
	\label{tab:judge-q0q1}
\end{table}

\begin{table}[p]
	\centering
	\footnotesize
	\setlength{\tabcolsep}{3pt}
	\resizebox{\textwidth}{!}{
		\begin{tabular}{ccl r ccccc}
			\toprule
			Subject & Judge & Model & $\theta$
			& $\hat\theta_{\mathrm{naive}}$
			& $\hat\theta_{\mathrm{RG}}$
			& $\hat\theta_{\mathrm{RG},2\times}$
			& $\hat\theta_{\mathrm{PPI{+}{+}}}$
			& $\hat\theta_{\mathrm{PPI{+}{+}},2\times}$ \\
			\midrule
			
			\multirow{12}{*}{\rotatebox[origin=c]{90}{Math}}
			& \multirow{6}{*}{\rotatebox[origin=c]{90}{Mistral Large}}
			& \multirow{2}{*}{Gemma\,3} & \multirow{2}{*}{0.738}
			& 0.776 & 0.668 & 0.720 & 0.742 & 0.742 \\
			&&&& $[0.749,\;0.802]$ & $[0.571,\;0.745]$ & $[0.589,\;0.826]$ & $[0.677,\;0.804]$ & $[0.702,\;0.781]$ \\[4pt]
			
			&& \multirow{2}{*}{Qwen\,2.5} & \multirow{2}{*}{0.735}
			& 0.822 & 0.756 & 0.756 & 0.752 & 0.749 \\
			&&&& $[0.798,\;0.845]$ & $[0.678,\;0.825]$ & $[0.678,\;0.825]$ & $[0.695,\;0.807]$ & $[0.712,\;0.785]$ \\[4pt]
			
			&& \multirow{2}{*}{diff} & \multirow{2}{*}{$+$0.003}
			& $-$0.046 & $-$0.089 & $-$0.037 & $-$0.009 & $-$0.007 \\
			&&&& $[-0.078,\;-0.016]$ & $[-0.156,\;-0.030]$ & $[-0.176,\;0.081]$ & $[-0.084,\;0.059]$ & $[-0.050,\;0.034]$ \\
			\cmidrule(l){2-9}
			
			& \multirow{6}{*}{\rotatebox[origin=c]{90}{Gemma\,4\,31B}}
			& \multirow{2}{*}{Gemma\,3} & \multirow{2}{*}{0.738}
			& 0.710 & 0.690 & 0.739 & 0.741 & 0.743 \\
			&&&& $[0.679,\;0.741]$ & $[0.618,\;0.753]$ & $[0.666,\;0.808]$ & $[0.683,\;0.799]$ & $[0.707,\;0.779]$ \\[4pt]
			
			&& \multirow{2}{*}{Qwen\,2.5} & \multirow{2}{*}{0.735}
			& 0.749 & 0.747 & 0.747 & 0.750 & 0.747 \\
			&&&& $[0.720,\;0.777]$ & $[0.681,\;0.808]$ & $[0.681,\;0.808]$ & $[0.694,\;0.805]$ & $[0.712,\;0.781]$ \\[4pt]
			
			&& \multirow{2}{*}{diff} & \multirow{2}{*}{$+$0.003}
			& $-$0.038 & $-$0.057 & $-$0.008 & $-$0.008 & $-$0.004 \\
			&&&& $[-0.070,\;-0.007]$ & $[-0.106,\;-0.010]$ & $[-0.086,\;0.072]$ & $[-0.083,\;0.060]$ & $[-0.044,\;0.037]$ \\
			\midrule
			
			\multirow{12}{*}{\rotatebox[origin=c]{90}{Biology}}
			& \multirow{6}{*}{\rotatebox[origin=c]{90}{Mistral Large}}
			& \multirow{2}{*}{Gemma\,3} & \multirow{2}{*}{0.766}
			& 0.699 & 0.244 & 0.575 & 0.807 & 0.806 \\
			&&&& $[0.659,\;0.740]$ & $[0.000,\;0.444]$ & $[0.000,\;1.000]$ & $[0.717,\;0.890]$ & $[0.755,\;0.855]$ \\[4pt]
			
			&& \multirow{2}{*}{Qwen\,2.5} & \multirow{2}{*}{0.718}
			& 0.826 & 0.577 & 0.577 & 0.658 & 0.665 \\
			&&&& $[0.795,\;0.857]$ & $[0.381,\;0.700]$ & $[0.381,\;0.700]$ & $[0.573,\;0.743]$ & $[0.611,\;0.715]$ \\[4pt]
			
			&& \multirow{2}{*}{diff} & \multirow{2}{*}{$+$0.048}
			& $-$0.127 & $-$0.330 & $-$0.002 & $+$0.149 & $+$0.142 \\
			&&&& $[-0.173,\;-0.079]$ & $[-0.493,\;-0.194]$ & $[-0.635,\;0.528]$ & $[0.059,\;0.237]$ & $[0.090,\;0.195]$ \\
			\cmidrule(l){2-9}
			
			& \multirow{6}{*}{\rotatebox[origin=c]{90}{Gemma\,4\,31B}}
			& \multirow{2}{*}{Gemma\,3} & \multirow{2}{*}{0.766}
			& 0.667 & 0.417 & 0.570 & 0.805 & 0.804 \\
			&&&& $[0.623,\;0.711]$ & $[0.235,\;0.546]$ & $[0.000,\;1.000]$ & $[0.715,\;0.886]$ & $[0.752,\;0.853]$ \\[4pt]
			
			&& \multirow{2}{*}{Qwen\,2.5} & \multirow{2}{*}{0.718}
			& 0.782 & 0.634 & 0.634 & 0.669 & 0.671 \\
			&&&& $[0.745,\;0.818]$ & $[0.508,\;0.730]$ & $[0.508,\;0.730]$ & $[0.586,\;0.750]$ & $[0.619,\;0.720]$ \\[4pt]
			
			&& \multirow{2}{*}{diff} & \multirow{2}{*}{$+$0.048}
			& $-$0.115 & $-$0.218 & $-$0.066 & $+$0.137 & $+$0.133 \\
			&&&& $[-0.165,\;-0.063]$ & $[-0.341,\;-0.117]$ & $[-0.641,\;0.346]$ & $[0.042,\;0.224]$ & $[0.082,\;0.185]$ \\
			\midrule
			
			\multirow{12}{*}{\rotatebox[origin=c]{90}{Law}}
			& \multirow{6}{*}{\rotatebox[origin=c]{90}{Mistral Large}}
			& \multirow{2}{*}{Gemma\,3} & \multirow{2}{*}{0.334}
			& 0.723 & 0.351 & 0.019 & 0.367 & 0.370 \\
			&&&& $[0.694,\;0.752]$ & $[0.000,\;0.650]$ & $[0.000,\;0.378]$ & $[0.292,\;0.448]$ & $[0.322,\;0.421]$ \\[4pt]
			
			&& \multirow{2}{*}{Qwen\,2.5} & \multirow{2}{*}{0.308}
			& 0.698 & 0.216 & 0.216 & 0.356 & 0.359 \\
			&&&& $[0.669,\;0.728]$ & $[0.000,\;0.503]$ & $[0.000,\;0.503]$ & $[0.281,\;0.434]$ & $[0.310,\;0.408]$ \\[4pt]
			
			&& \multirow{2}{*}{diff} & \multirow{2}{*}{$+$0.026}
			& $+$0.025 & $+$0.126 & $-$0.130 & $+$0.011 & $+$0.011 \\
			&&&& $[-0.014,\;+0.064]$ & $[-0.071,\;+0.374]$ & $[-0.457,\;+0.245]$ & $[-0.077,\;+0.103]$ & $[-0.045,\;+0.068]$ \\
			\cmidrule(l){2-9}
			
			& \multirow{6}{*}{\rotatebox[origin=c]{90}{Gemma\,4\,31B}}
			& \multirow{2}{*}{Gemma\,3} & \multirow{2}{*}{0.334}
			& 0.480 & 0.238 & 0.263 & 0.361 & 0.365 \\
			&&&& $[0.443,\;0.515]$ & $[0.102,\;0.358]$ & $[0.089,\;0.414]$ & $[0.289,\;0.438]$ & $[0.320,\;0.413]$ \\[4pt]
			
			&& \multirow{2}{*}{Qwen\,2.5} & \multirow{2}{*}{0.308}
			& 0.519 & 0.319 & 0.319 & 0.352 & 0.357 \\
			&&&& $[0.482,\;0.556]$ & $[0.190,\;0.437]$ & $[0.190,\;0.437]$ & $[0.283,\;0.423]$ & $[0.311,\;0.401]$ \\[4pt]
			
			&& \multirow{2}{*}{diff} & \multirow{2}{*}{$+$0.026}
			& $-$0.040 & $-$0.081 & $-$0.056 & $+$0.010 & $+$0.009 \\
			&&&& $[-0.083,\;+0.005]$ & $[-0.177,\;+0.011]$ & $[-0.245,\;+0.129]$ & $[-0.076,\;+0.096]$ & $[-0.045,\;+0.064]$ \\
			\midrule
			
			\multirow{12}{*}{\rotatebox[origin=c]{90}{Engineering}}
			& \multirow{6}{*}{\rotatebox[origin=c]{90}{Mistral Large}}
			& \multirow{2}{*}{Gemma\,3} & \multirow{2}{*}{0.393}
			& 0.476 & 0.610 & 0.331 & 0.427 & 0.425 \\
			&&&& $[0.443,\;0.510]$ & $[0.474,\;0.781]$ & $[0.128,\;0.512]$ & $[0.330,\;0.515]$ & $[0.373,\;0.476]$ \\[4pt]
			
			&& \multirow{2}{*}{Qwen\,2.5} & \multirow{2}{*}{0.380}
			& 0.383 & 0.402 & 0.402 & 0.382 & 0.377 \\
			&&&& $[0.351,\;0.417]$ & $[0.275,\;0.541]$ & $[0.275,\;0.541]$ & $[0.304,\;0.464]$ & $[0.328,\;0.428]$ \\[4pt]
			
			&& \multirow{2}{*}{diff} & \multirow{2}{*}{$+$0.012}
			& $+$0.093 & $+$0.208 & $-$0.073 & $+$0.044 & $+$0.047 \\
			&&&& $[+0.056,\;+0.128]$ & $[+0.122,\;+0.319]$ & $[-0.291,\;+0.122]$ & $[-0.049,\;+0.133]$ & $[-0.010,\;+0.102]$ \\
			\cmidrule(l){2-9}
			
			& \multirow{6}{*}{\rotatebox[origin=c]{90}{Gemma\,4\,31B}}
			& \multirow{2}{*}{Gemma\,3} & \multirow{2}{*}{0.391}
			& 0.242 & 0.367 & 0.364 & 0.429 & 0.424 \\
			&&&& $[0.208,\;0.276]$ & $[0.253,\;0.506]$ & $[0.219,\;0.521]$ & $[0.328,\;0.521]$ & $[0.373,\;0.475]$ \\[4pt]
			
			&& \multirow{2}{*}{Qwen\,2.5} & \multirow{2}{*}{0.379}
			& 0.230 & 0.338 & 0.338 & 0.373 & 0.369 \\
			&&&& $[0.197,\;0.262]$ & $[0.226,\;0.471]$ & $[0.226,\;0.471]$ & $[0.298,\;0.452]$ & $[0.321,\;0.420]$ \\[4pt]
			
			&& \multirow{2}{*}{diff} & \multirow{2}{*}{$+$0.012}
			& $+$0.012 & $+$0.029 & $+$0.023 & $+$0.055 & $+$0.054 \\
			&&&& $[-0.019,\;+0.043]$ & $[-0.046,\;+0.108]$ & $[-0.118,\;+0.174]$ & $[-0.039,\;+0.148]$ & $[-0.001,\;+0.110]$ \\
			\bottomrule
		\end{tabular}
	}
\caption{Judge-based estimates of model accuracy and model differences on MMLU-Pro math, biology, law, and engineering. $\theta$ denotes true test-set accuracy. ``diff'' rows show Gemma\,3 minus Qwen\,2.5. RG uses shared calibration; RG\,$2\times$ uses model-specific calibration with the RG formula; PPI\texttt{++} variants use model-specific calibration.}
	\label{tab:judge-theta-estimates}
\end{table}

\end{document}